\pdfoutput=1

\documentclass[11pt]{article}

\usepackage[]{EMNLP2022}

\usepackage{times}
\usepackage{latexsym}

\usepackage[T1]{fontenc}

\usepackage[utf8]{inputenc}

\usepackage{microtype}

\usepackage{inconsolata}

\usepackage{makecell}
\usepackage{enumitem}
\usepackage{xcolor,soul} 
\usepackage{booktabs}
\usepackage{enumitem}
\usepackage{float}
\usepackage{color}
\usepackage{pgf,tikz,pgfplots}
\usepackage{colortbl}
\restylefloat{table}
\usepackage{subcaption}
\usepackage{amsmath}
\usepackage{amssymb}
\usepackage{pifont}%
\usepackage{mathtools}
\usepackage{subcaption}
\usepackage{xspace}
\definecolor{beaublue}{rgb}{0.74, 0.83, 0.9}
\DeclareRobustCommand{\hlcyan}[1]{{\sethlcolor{beaublue}\hl{#1}}}
\usepackage{multirow}
\usepackage{makecell}
\newcommand{\cmark}{\ding{51}}%
\newcommand{\xmark}{\ding{55}}%

\newcommand\robotemoji{\includegraphics[width=1em]{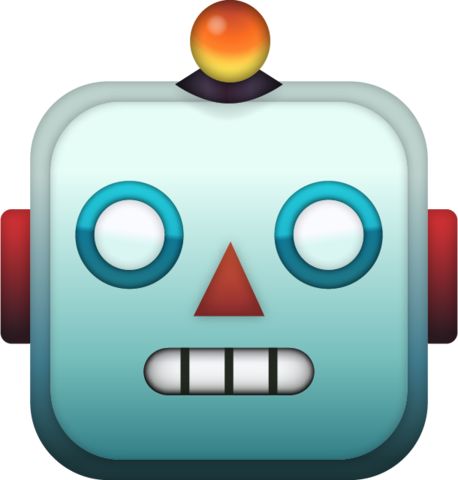}\xspace}
\newcommand\personemoji{\raisebox{-2pt}{\includegraphics[width=1.3em]{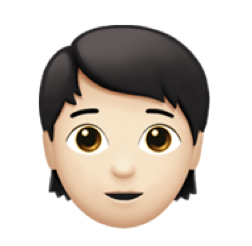}}\xspace}

\usepackage{fontawesome}

\makeatletter
\def\thanks#1{\protected@xdef\@thanks{\@thanks
        \protect\footnotetext{#1}}}
\makeatother

\usepackage{times}
\usepackage{latexsym}
\usepackage{bbold}

\newcommand{\datasetsize}{14,182 }
\usepackage[T1]{fontenc}
\usepackage{linguex}
\usepackage[utf8]{inputenc}

\usepackage{microtype}

\usepackage{tikz}
\usepackage{tabularx}
\usepackage{amsmath}
\usepackage{amsthm}
\usepackage{amssymb}
\usepackage{paralist}
\usepackage[capitalise]{cleveref}
\usepackage{booktabs}
\usepackage{todonotes}
\usepackage{paralist}
\usepackage{xspace}
\usepackage{colortbl}

\newcommand{\draftonly}[1]{#1}
\newcommand{\draftcomment}[1]{\draftonly{#1}}

\newcommand\sect[1]{\S\ref{#1}}

\newif\ifcomments
\ifcomments
    \providecommand{\ana}[1]{\draftcomment{\textcolor{blue}{\small [Ana: #1]}}}
    \providecommand{\matt}[1]{\draftcomment{\textcolor{red}{\small [Matt: #1]}}}
    \providecommand{\lasha}[1]{\draftcomment{\textcolor{magenta}{\small [Lasha: #1]}}}
\else
    \providecommand{\matt}[1]{}
    \providecommand{\ana}[1]{}
    \providecommand{\lasha}[1]{}
\fi

\newcommand{\datasetname}{\textsc{CondaQA}\xspace}

\newcommand{\uqalarge}{\textsc{UnifiedQA-v2-large}\xspace}
\newcommand{\uqathreeb}{\textsc{UnifiedQA-v2-3B}\xspace}
\usepackage[normalem]{ulem}
\newcommand{\change}[1]{\draftcomment{\textcolor{black}{#1}}}
\title{\datasetname{}: A \emph{Contrastive} Reading Comprehension Dataset for\\Reasoning about Negation}

\author{Abhilasha Ravichander$^*$ \\
Carnegie Mellon University \\
  \texttt{aravicha@cs.cmu.edu} \\ \And
  Matt Gardner\\
  Microsoft Semantic Machines \\
  \texttt{mattgardner@microsoft.com} \\\And
  Ana Marasovi\'{c}$^*$ \\
  University of Utah \\
  \texttt{ana.marasovic@utah.edu} \\}

\begin{document}
\maketitle
\begingroup\def\thefootnote{*}\footnotetext{Work undertaken while Abhilasha Ravichander and Ana Marasovi\'{c} were at the Allen Institute for AI.}\endgroup
\begin{abstract}
The full power of human language-based communication cannot be realized without negation. All human languages have some form of negation. Despite this, negation remains a challenging phenomenon for current natural language understanding systems. To facilitate the future development of models that can process negation effectively, we present \datasetname, the first English reading comprehension dataset which requires reasoning about the implications of negated statements in paragraphs. We collect paragraphs with diverse negation cues, then have crowdworkers ask questions about the \emph{implications} of the negated statement in the passage.  We also have workers make three kinds of edits to the passage---paraphrasing the negated statement, changing the scope of the negation, and reversing the negation---resulting in clusters of question-answer pairs that are difficult for models to answer with spurious shortcuts.
\datasetname features \datasetsize question-answer pairs with over 200 unique negation cues and is challenging for current state-of-the-art models. %
\change{The best performing model on \datasetname (\uqathreeb) achieves only 42\% on our consistency metric, well below human performance which is 81\%.} %
We release our dataset, along with fully-finetuned, few-shot, \change{and zero-shot evaluations}, to facilitate the development of future NLP methods that work on negated language.

\end{abstract}



\begin{figure*}[!tb]

    \includegraphics[width=\textwidth]{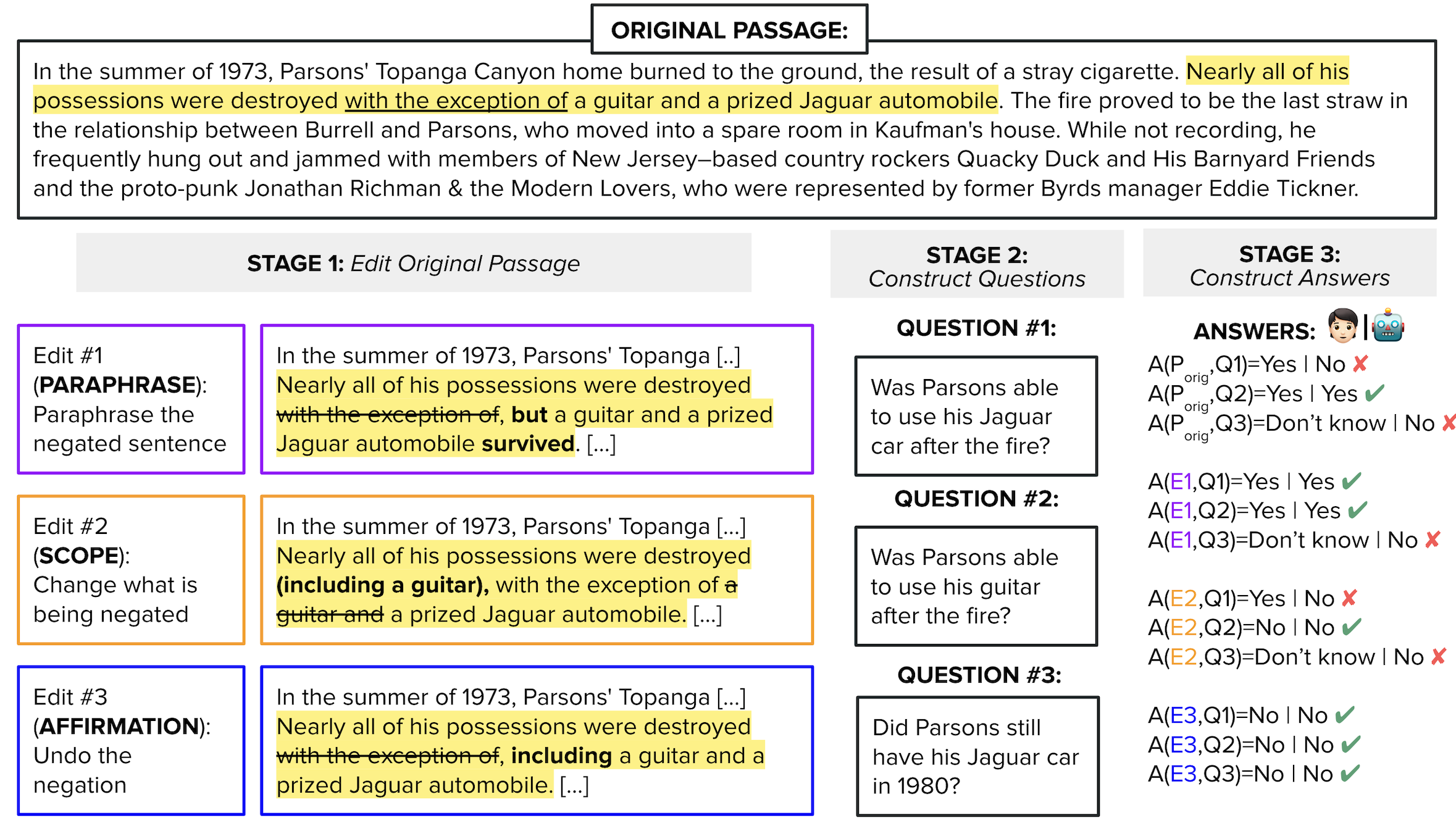}
  \caption{\datasetname three-stage collection procedure. The original passage is selected by a crowdworker from a given set of 10 passages. \personemoji Gold answers given by crowdworkers; \robotemoji Answers predicted by  \faMagic InstructGPT (\texttt{text-davinci-002}) prompted with 8 shots. See \sect{sec:condaqa_data_collection} for more details about each stage.}
 \label{ref:data-exp}
\end{figure*}
\section{Introduction}
\label{sec:intro}

Negation is fundamental to human communication. It is a phenomenon of semantic opposition, relating one expression to another whose meaning is in some way opposed. Negation supports key properties of human linguistic systems such as contradiction and denial~\cite{horn1989natural}. 

Despite the prevalence of negation, processing it effectively continues to elude models. %
Here are just a few of the many recently reported failures: 
``The model [BERT-Large trained on SQuAD] does not seem capable of handling...simple examples of negation'' \cite{ribeiro-etal-2020-beyond}. ``We find that indeed the presence of negation can significantly impact downstream quality [of machine translation systems]'' \cite{hossain-etal-2020-non}. ``State-of-the-art models answer questions from the VQA...correctly, but struggle when asked a logical composition including negation'' \cite{gokhale2020vqa}. \emph{How can NLU systems meet this long-standing challenge?} 

To facilitate systems that can process negation effectively, it is crucial to have high-quality evaluations that accurately  measure models' competency at processing and understanding negation.  %
In this work, we take a step toward this goal by contributing the first large-scale reading comprehension dataset, \datasetname, focused on reasoning about negated statements in language.\footnote{\textbf{CO}ntrastively-annotated \textbf{N}egation \textbf{DA}taset of \textbf{Q}uestion-\textbf{A}nswer pairs} 

\change{The three-stage annotation process we develop to construct \datasetname is illustrated in Fig.\ \ref{ref:data-exp}.} 
We first collect passages from English Wikipedia that contain negation cues, including single- and multi-word negation phrases, as well as affixal negation. 
In the first stage, crowdworkers make three types of modifications to the original passage: (1) they paraphrase the negated statement, (2) they modify the scope of the negated statement (while retaining the negation cue), and (3) they undo the negation. 
In the second stage, we instruct crowdworkers to ask challenging questions about the \emph{implications} of the negated statement. 
The crowdworkers then answer the questions they wrote previously for the original and edited passages. 

This process resulted in a dataset of \datasetsize questions, covering a variety of negation cue types and over 200 unique negation cues, as well as a \emph{contrastive} dataset, with passages that are lexically similar to each other but that may induce different answers for the same questions. To perform well on \datasetname, models must be able to reason about the implications of negated statements in text. In addition to accuracy, the contrastive nature of \datasetname enables us to measure the \emph{consistency} of models---i.e., the extent to which models make correct predictions on closely-related inputs. 

\change{We extensively benchmark baseline models on \datasetname in three training data regimes: using all training examples, using only a small fraction (few-shot), or not using any examples (zero-shot). %
We show that \datasetname is challenging for current models. Finetuning \textsc{Unified-QA-3B}~\cite{khashabi2022unifiedqa}---which was trained on 20 QA datasets---on \datasetname, achieves the best result of \change{73.26\%} compared to human accuracy of 91.49\%. %
Further, we find that models are largely inconsistent; the best model achieves a consistency score of only \change{42.18\%} (40\% below human consistency). }
This very low consistency score demonstrates that handling negation phenomena is still a major unresolved issue in NLP, along with sensitivity to contrasting data more generally. %
\change{The dataset and baselines are available at \url{https://github.com/AbhilashaRavichander/CondaQA}.}

\section{\datasetname Data Collection}
\label{sec:condaqa_data_collection}
This section describes our goals in constructing \datasetname and our data collection procedure.

\paragraph{Design Considerations} Our goal is to evaluate models on their ability to process the contextual implications of negation. We have the following four desiderata for our question-answering dataset:

\begin{compactenum}
\item The dataset should include a wide variety of ways negation can be expressed.
\item Questions should be targeted towards the \emph{implications} of a negated statement, rather than the factual content of what was or wasn't negated, to remove common sources of spurious cues in QA datasets~\cite{kaushik-lipton-2018-much,naik-etal-2018-stress,mccoy-etal-2019-right}.
\item The dataset should feature contrastive groups: passages that are closely-related, but that may admit different answers to questions, in order to reduce models’ reliance on potential spurious cues in the data and to enable more robust evaluation~\cite{kaushik2019learning,gardner-etal-2020-evaluating}.
\item Questions should probe the extent to which models are sensitive to how the negation is expressed. In order to do this, there should be contrasting passages that differ only in their negation cue or its scope.
\end{compactenum}

\paragraph{Dataset Construction Overview} We generate questions through a process that consists of the following steps, as shown in Figure \ref{ref:data-exp}:
\begin{compactenum}
    \item We extract passages from Wikipedia which contain negated phrases. 
    \item We show ten passages to crowdworkers, and allow them to choose a passage they would like to work on. 
    \item Crowdworkers make three kinds of edits to the passage: (i) paraphrasing the negated statement, (ii) changing the scope of the negation, (iii) rewriting the passage to include an affirmative statement in place of the negated statement. For all three kinds of edits, the crowdworkers modified the passage as appropriate for internal consistency.
    \item Crowdworkers ask questions that target the implications of a negated statement in the passage, taking passage context into account.
    \item Crowdworkers provide answers to the constructed questions for the Wikipedia passage, as well as the three edited passages.
\end{compactenum}

Further, we validate the development and test portions of \datasetname to ensure their quality.

\paragraph{Passage Selection} We extract passages from a July 2021 version of Wikipedia that contain either single-word negation cues (e.g., `no') or multi-word negation cues  (e.g., `in the absence of'). %
We use negation cues from \cite{morante2011annotation,van-son-etal-2016-building} as a starting point which we extend. %
Our single-word negation cues include affixal negation cues (e.g., `\emph{il}-legal'), and span several grammatical categories including:

\begin{compactenum}
\item \textbf{Verbs}: In this novel, he took pains to \hlcyan{avoid} the scientific impossibilities which had bothered some readers of the "Skylark" novels.
\item \textbf{Nouns}:  In the \hlcyan{absence} of oxygen, the citric acid cycle ceases.
\item  \textbf{Adjectives}: Turning the club over to managers, later revealed to be honest people, still left Wills in desperate financial straits with heavy debts to the \hlcyan{dishonest} IRS for 
taxes.
\item  \textbf{Adverbs}: Nasheed reportedly resigned \hlcyan{involuntarily} to forestall an escalation of violence;
\item  \textbf{Prepositions}: Nearly half a century later, after Fort Laramie had been built \hlcyan{without} permission on Lakota land. 
\item  \textbf{Pronouns}:  I mean, \hlcyan{nobody} retires anymore.
\item  \textbf{Complementizers}: Leave the door ajar, \hlcyan{lest} any latecomers should find themselves shut out. 
\item  \textbf{Conjunctions}: Virginia has no `pocket veto' and bills will become law if the governor chooses to neither approve \hlcyan{nor} veto legislation.
\item  \textbf{Particles}: Botham did \hlcyan{not} bat again. 
\end{compactenum}


\begin{table}[t]
\centering
\small
\begin{tabular}{lccc}
\toprule
                                            & Train & Dev & Test \\
\midrule

\# Passages                                 &  474     & 115    &   700  \\
\midrule
Average passage length                     &  130.02     &  131.24   & 131.0    \\
\midrule
Negated statement length            &   28.12    &    29.96 &  28.0   \\
\midrule
\# Unique negation cues            & 134       & 62  & 171    \\
\midrule
\# Unseen negation cues            & -       &  18 & 75    \\
\midrule
\# Questions                                &  5832     &   1110  &  7240   \\
\midrule
Average Question Length                     &  24.2     &  26.38   &   24.35  \\
\midrule
\# Questions w/ \textgreater{}20 tokens &   2836    &  650   & 3616   \\
\midrule
\# Distinct question words                     &  6045     &  2235   &  7603   \\
\bottomrule
\end{tabular}
\caption{Dataset statistics of \datasetname. Passage statistics are computed on Wiki passages but not on edits.  } 
\label{ref:datasetstats}
\end{table}


\paragraph{Crowdworker Recruitment} We use the Crowdaq platform~\cite{ning-etal-2020-easy} to recruit a small pool of qualified workers to contribute to \datasetname. We provide instructions, a tutorial and a qualification task. Workers were asked to read the instructions, and optionally to also go through the tutorial. Workers then took a qualification exam which consisted of 12 multiple-choice questions that evaluated comprehension of the instructions. %
We recruit crowdworkers who answer $>$70\% of the questions correctly for the next stage of the  dataset construction task. In total, 36 crowdworkers contributed to \datasetname. We paid 8 USD/HIT, which could on average be completed in less than 30 minutes. Each HIT consisted of choosing a passage, making edits to the passage, creating questions, and answering those questions. 

\paragraph{Contrastive Dataset Construction}
We use Amazon Mechanical Turk 
to crowdsource question-answer pairs about negated statements. Each question is asked in the context of a negated statement in a Wikipedia passage.  

In the first stage of the task, we show crowdworkers ten selected passages of approximately the same length and let them choose which to work on. This allows crowdworkers the flexibility to choose passages which are easy to understand, as well as to choose passages which are conducive to making contrastive edits (for example, it may be difficult to reverse the negation in a passage about `Gödel's \emph{incompleteness} theorems'). 

After selecting a passage, crowdworkers make three kinds of edits to the original Wikipedia passage (Fig.\ \ref{ref:data-exp}): (1) they rewrite the negated statement such that the sentence's meaning is preserved (\textsc{Paraphrase Edit}); (2) they rewrite the negated statement, changing the scope of the negation (\textsc{Scope Edit}); and (3) they reverse the negated event (\textsc{Affirmative Edit}).  We ask crowdworkers to additionally make edits outside of the negated statement where necessary to ensure that the passage remains internally consistent.

In the second stage of the task, the crowdworker asks at least three questions about the implications of the negated statement in the original Wikipedia passage. We encourage the construction of good questions about implications by providing several examples of such questions, as well as by sending bonuses to creative crowdworkers, ranging from 10\$-15\$.  Crowdworkers can group these questions, to indicate questions that are very similar to each other, but admit different answers.

In the final stage of this task, crowdworkers provide answers to the questions, in context of the Wikipedia passages as well as for the three edited passages. The answers to the questions are required to be either Yes/No/Don't Know, a span in the question, or a span in the passage. Following best practices for crowdsourcing protocols described in the literature~\cite{nangia-etal-2021-ingredients}, we provide personalized feedback to each crowdworker based on their previous round of submissions, describing where their submission was incorrect, why their submission was incorrect, and what they could have submitted instead. In all, we provide over 15 iterations of expert feedback on the annotations. We collect this data over a period of $\sim$seven months.

\paragraph{Data Cleaning and Validation}
In order to estimate human performance, and to construct a high-quality evaluation with fewer ambiguous examples, we have five verifiers provide answers for each question in the development and test sets. Crowdworkers were given passages, as well as the passage edits and questions contributed in the previous stage of our task. In each HIT, crowdworkers answered 60 questions in total, spanning five passage sets. We found there was substantial inter-annotator agreement; for the test set we observed a Fleiss' $\kappa$ of 63.27 for examples whose answers are Yes/No/Don't know (97\% of examples), 62.75 when answers are a span in the question (2\% of examples), and 48.54 when answers were indicated to be a span in the passage (1\% of examples). We only retain examples in the test and development sets where \emph{at least four annotators} agreed on the answer. However, since this procedure results in few questions with `don't know' as the answer, we include an additional stage where we (the authors) manually verify and include questions where `don't know' was the answer provided by the question author. As a result, we discard 1,160 instances from the test set, and 270 from the development set. 

\begin{figure}[tb]

    \includegraphics[width=\columnwidth]{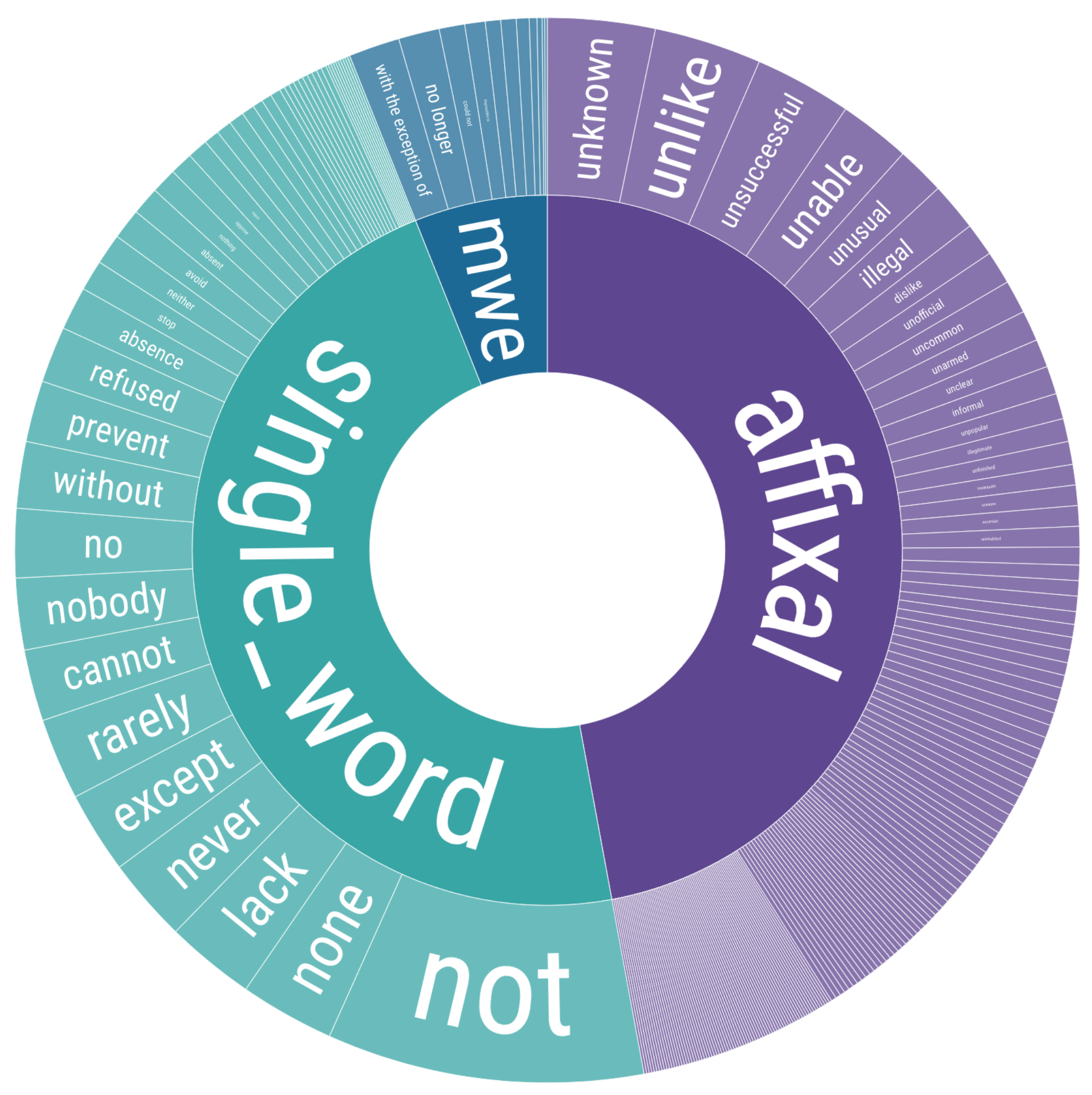}

  \caption{Distribution of negation cues in \datasetname. Inner circle represents distribution of negation cue types by frequency and the outer circle represents cues.}
 \label{ref:negcue-dist}
\end{figure}

\begin{table*}[!h]
\centering
\small
\begin{tabular}{p{0.08\textwidth}p{0.35\textwidth}p{0.25\textwidth}p{0.05\textwidth}p{0.15\textwidth}} \toprule
\textbf{Reasoning Type}     & \textbf{Passage Snippet} & \textbf{Question} & \textbf{Answer} & \textbf{Explanation} \\ \toprule
\emph{Social Norms}  (10\%)       &  \textcolor{blue}{On October 8, 1883, the US patent office ruled that Edison's patent was based on the work of William E. Sawyer and was, therefore, \textbf{invalid} .} Litigation continued for nearly six years. In 1885, Latimer switched camps and started working with Edison.         &  From the information given in the passage, would you say that coincidence is the most charitable explanation for what was essentially the same innovation, in much the same way that Newton and Leibniz seemingly discovered calculus independently, without knowing of the other's progress?       &  YES   & Plagarism is frowned upon in society, more so than accidentally reaching the same conclusions as someone else.      \\ 
\midrule
\emph{Psychology} (9\%)      &    [...] Disraeli later romanticised his origins, claiming his father's family was of grand Iberian and Venetian descent; in fact Isaac's family was of no great distinction [...] \textcolor{blue}{ Historians differ on Disraeli's motives for rewriting his family history: [...] Sarah Bradford believes ``his  \textbf{dislike}  of the commonplace would not allow him to accept the facts of his birth as being as middle-class and undramatic as they really were''. }        &   Would Disraeli have been flattered by a biography that explored his middle class upbringing, according to Bradford?      & NO   &  A person such as Disraeli who wants to project a grandiose image of themselves is likely to be unhappy when people discuss mundane aspects about his upbringing.   \\ \midrule
\multirow{2}{2cm}{\emph{Cause and Effect}  (7\%)  }         &     
Oil produced from palm fruit is called `red palm oil' or just `palm oil'... \textcolor{blue}{ In its \textbf{unprocessed}  state, red palm oil has an intense deep red color because of its abundant carotene content.} [...]

        &   Would a consumer who was primarily interested in the eye-health benefits of carotenes and lycopene want to shop for palm oils by their color, rather than listening to marketing slogans such as ``extra virgin'' or ``minimally processed''?   & YES & A high carotene content causes a deep red color, so a person searching for things with high carotene content can look at their color.  \\ \bottomrule    
\end{tabular}
\caption{Examples of questions that target the implications of negated statements in \datasetname and reasoning steps to correctly answer the questions.  Negated statements are in \textcolor{blue}{blue}.  Categories inspired by \newcite{lobue-yates-2011-types}.  Expanded analysis is shown in the Appendix (Table \ref{ref:qreasoning_types}).}
\label{ref:qreasoning_types_partial}
\end{table*}

\section{\datasetname Data Analysis}
In this section, we provide an analysis of the passages, questions, edits, and answers in \datasetname. Descriptive statistics are provided in Table \ref{ref:datasetstats}. 

\paragraph{Negation Cues} Negation is expressed in many complex and varied ways in language~\cite{horn1989natural}. %
To characterize the distribution of types of negated statements in \datasetname, we analyze the negation cues in Wikipedia passages that annotators could select. %
Figures \ref{ref:negcue-dist} and \ref{ref:negcue-grammar} (Appendix) show that the distribution over these cues and their grammatical roles is considerably diverse. %
Moreover, there are 219 unique cues in \datasetname and 75 novel cues in the test set that are unseen in the training data. %
This is a substantially wider range of negation cues than what is included in prior work; see Appendix \ref{ref:priorwork} for a detailed comparison.

\paragraph{Commonsense inferences} We assess commonsense inferences types required to answer \datasetname questions. We sample 100 questions from the test set and manually annotate the dimensions of commonsense reasoning required to answer them. Table~\ref{ref:qreasoning_types_partial} shows some of these reasoning types (the full version in Table \ref{ref:qreasoning_types} in the Appendix).

\definecolor{light-gray}{gray}{0.94}
\begin{table*}[!tb]
         \centering
\footnotesize
\begin{tabular}{p{0.1\textwidth}p{0.8\textwidth} }
  \toprule
\textbf{Revision Strategy} & \textbf{Edited Passage} \\
  \midrule
    \multicolumn{2}{c}{\cellcolor{light-gray} \textsc{Paraphrase Edit}} \\
\multirow{4}{2cm}{\emph{Complement substitution}}  & Though Philby claimed publicly in January 1988 that he did not regret his decisions and that \sout{he missed nothing about England \textcolor{blue}{except} }\textcolor{red}{the only things he missed about England were} some friends, Colman's mustard, and Lea \& Perrins Worcestershire sauce... \\
\arrayrulecolor{black!20}\midrule
\emph{Synonym substitution}  & Local tetanus is \sout{an \textcolor{blue}{uncommon}}\textcolor{red}{a rare} form of the disease and it causes persistent contractions of muscles in the same area of the sufferer's body as where the original injury was made.\\
\arrayrulecolor{black!20}\midrule
\emph{Antonym substitution}  & The population of the Thirteen States was \sout{\textcolor{blue}{not} homogeneous}  \textcolor{red}{heterogeneous} in political views and attitudes. \\
\arrayrulecolor{black!20}\midrule
\multirow{2}{*}{\emph{Ellipsis}} & \sout{Sunni scholars put trust in narrators such as Aisha, whom Shia  \textcolor{blue}{reject}}\textcolor{red}{While the Shia tend to reject} narrators such as Aisha, \textcolor{red}{Sunni scholars tend to trust them.}\\
    \multicolumn{2}{c}{\cellcolor{light-gray} \textsc{Scope Edit}} \\
\emph{Complement inversion} &  \sout{Sunni}\textcolor{red}{Shia} scholars put trust in narrators such as Aisha, whom \sout{Shia}\textcolor{red}{Sunni} \textcolor{blue}{reject}.\\
\arrayrulecolor{black!20}\midrule
\emph{Superset-subset} & During the coronavirus outbreak of 2020, alcohol sales\sout{, and even the} \textcolor{red}{were made} \textcolor{blue}{illegal}, \textcolor{red}{but the} transportation of alcohol outside of one's home\sout{, was made \textcolor{blue}{illegal}} \textcolor{red}{remained legal}.\\
\arrayrulecolor{black!20}\midrule
\emph{Temporal shift} & As the new Emperor \textcolor{blue}{could not} exert his constitutional powers \sout{until}\textcolor{red}{once} he came of age, a regency was set up by the National Assembly. \\
\arrayrulecolor{black!20}\midrule
\multirow{3}{*}{\emph{Veridicality}} &  \textcolor{red}{Contrary to assumptions that he was}  \textcolor{blue}{illiterate}, on arrival he was given aptitude tests which determined that \sout{he was \textcolor{blue}{illiterate}}\textcolor{red}{not only could he read the questions and respond in writing}, but \textcolor{red}{he also} had an above-average IQ of 109. \\
 
\arrayrulecolor{black}\bottomrule
  \end{tabular}

\caption{Examples of revision strategies employed by crowdworkers for paraphrase and scope edits. Categories for paraphrases are inspired by \newcite{bhagat-hovy-2013-squibs}. The negation cue is shown in \textcolor{blue}{blue} and newly-inserted text is in \textcolor{red}{red}.  Expanded analysis is shown in the Appendix (Table \ref{tab:EditTypes}). }
\label{tab:EditTypes_partial}
\end{table*}


\paragraph{Editing Strategies} Recall that the passages with negated statements are sourced from Wikipedia and crowdworkers make three kinds of edits (Fig.\ \ref{ref:data-exp}).  
Through a qualitative analysis of the data, we identify commonly employed edit strategies  (Tables~\ref{tab:EditTypes_partial} and \ref{tab:EditTypes}). We also analyze to what extent edits cause an answer to change. We find that the affirmative edits change the answers of 77.7\% of questions from the original Wikipedia passage, and the scope edits change the answer of 70.6\% of questions.

\paragraph{Potential edit artifacts} Because we had crowdworkers edit Wikipedia paragraphs, a potential concern is that the edited text could be unnatural and give spurious cues to a model about the correct answer.  We ran two tests to try to quantify potential bias in this edited data.  First, we trained a BERT model~\cite{devlin-etal-2019-bert} to predict the edit type given just the passage.  The model performs only a little better than random chance (34.4\%), most of the improvement coming from the ability to sometimes detect affirmative edits (where the negation cue has been removed).  Second, we compared the perplexity of the original paragraphs to the perplexity of the edited paragraphs, according to the GPT language model~\cite{radford2018improving}, finding that they are largely similar.  Details for both of these experiments are in Appendix~\ref{ref:data-analysis-appendix}.

\section{Baseline Performance on \datasetname}
\label{ref:baselines_and_metrics}

We now evaluate state-of-the-art models' abilities to solve instances of \datasetname. 
We evaluate models that we train either on the entire \datasetname training data or few examples, as well as zero-shot models. %
We use two classes of metrics:

\paragraph{Accuracy} The percentage of predictions which match the ground truth answer. If the answer is a span, this metric measures whether the prediction matches the ground truth answer exactly.  
\paragraph{Group Consistency} \datasetname's dense annotations enable us to study model robustness through group consistency. We wish to measure whether a model correctly captures how the presence of negated phrases influences what can be inferred from a paragraph.  Measuring this requires varying (and sometimes removing) the negated phrases and seeing how the model responds (see Table \ref{fig:appendix_interpreting_results} in the Appendix); it is only by looking at consistency across these perturbations that we can tell whether a model understands the phenomena in question~\cite{gardner-etal-2020-evaluating}.  Specifically, for a group of minimally-different instances, consistency measures whether the prediction matches the ground truth answer for every element in that group. %
We consider two types of groups: (a) \emph{Question-level consistency}: each group is formed around a question and the answers to that question for the original Wikipedia passage, as well as the three edited passage instances (\textsc{All}), (b) \emph{Edit-level consistency}: each group is formed around a question, the answers to that question for the original Wikipedia passage, and only one of the edited passages (\textsc{Paraphrase consistency}, \textsc{Scope consistency}, and \textsc{Affirmative consistency}). To compute consistency, we use the 5,608 questions in the test set that have (passage, answer) pairs for all four edit types (excluding any question where at least one passage was removed during validation).

\subsection{Models and Controls}
The baseline models that we benchmark on \datasetname are listed in Table \ref{tab:ModelMainResults}. 
We categorize them based on whether they use (a) all of the training data we provide (full finetuned), (b) a small fraction of the available training data (few-shot), (c) no training data (zero-shot), and on (d) whether they measure dataset artifacts (controls).

For \textbf{full finetuning}, we train and evaluate three BERT-like models: BERT~\cite{devlin-etal-2019-bert}, RoBERTa~\cite{liu2019roberta}, and DeBERTa ~\cite{he2020deberta,he2021debertav3}, in addition to UnifiedQA-v2 \cite{khashabi2022unifiedqa}, a T5 variant~\cite{raffel2020exploring} that was further specialized for QA by training the model on 20 QA datasets. %
More information about these models is given in Appendix \ref{sec:appendix_supervised_models}. %
We study Base, Large, and 3B sizes of UnifiedQA-v2. %
Each fully-finetuned model is trained with 5 random seeds, and we report the average performance across seeds on the entire test set.

In the \textbf{few-shot} setting  with 8--9 shots, we evaluate UnifiedQA-v2-\{Base, Large, 3B\}~\cite{khashabi2022unifiedqa}, GPT-3~\cite[\texttt{davinci};][]{brown2020language}, and a version of InstructGPT$_\textit{orig}$ \cite{DBLP:journals/corr/abs-2203-02155} known as \texttt{text-davinci-002}; henceforth referred to as \faMagic InstructGPT. %
We additionally prompt \faMagic InstructGPT with chain of thoughts \cite[CoT;][]{DBLP:journals/corr/abs-2201-11903} as this should be beneficial for reasoning tasks. %
We do prompt-based finetuning of UnifiedQA-v2 (i.e., change its parameters) and in-context learning of the GPT models (i.e., we do not change their  parameters). Besides these models, in the \textbf{zero-shot} setting, we also evaluate UnifiedQA-v2-11B and FLAN-T5-11B \cite{Chung2022ScalingIL}, a T5 variant that was further trained with instruction finetuning and CoT data. %
Details of few- and zero-shot settings are given in Appendix \ref{sec:appendix_few_zero_shot_details}. %
Due to the cost of the OpenAI API and sensitivity of few-shot learning to the choice of few examples \cite{DBLP:conf/icml/ZhaoWFK021, logan-iv-etal-2022-cutting, DBLP:conf/nips/PerezKC21}, we evaluate few- and zero-shot models as follows. %
We split the train/test sets into five disjoint sets, sample 9 shots from each train subset, evaluate models on such five train-test splits, and report the average performance across them. 
On average each test split contains 1448 instances. 

We evaluate \textbf{heuristic} baselines to measure the extent to which models can use data artifacts to answer \datasetname questions. 
These baselines can answer questions correctly only if there is bias in the answer distribution given a question or other metadata since they do not get paragraphs. %
We train \uqalarge on just: (i) (question, answer) pairs, (ii) (question, edit type, answer) triples where the edit type denotes whether the passage was a paraphrase, scope edit, etc., and (iii) (question, negation cue, answer) triples. 
We find these baselines do little better than just answering ``No''.

\paragraph{Human Performance} We measure human performance on \datasetname development and test sets. %
Every question was answered by five crowdworkers. %
To evaluate human performance, we treat each answer to a question as the human prediction in turn, and compare it with the most frequent answer amongst the remaining answers. %
For questions where the gold answer was decided by experts (\S\ref{sec:condaqa_data_collection}), we treat each answer as the human prediction and compare it to the gold answer. %
Human accuracy is 91.94\%, with a consistency score of 81.58\%.

\definecolor{light-gray}{gray}{0.94}
\begin{table*}[!h]
         \centering
\small
\resizebox{\textwidth}{!}{
  \begin{tabular}
  {lcrrrrr}
  \toprule
\textbf{Model} & \textbf{\# Param} & \textbf{Accuracy} & \textbf{Consistency} & \makecell[r]{\textbf{Paraphrase}\\ \textbf{Consistency}}  &  \makecell[r]{\textbf{Scope}\\\textbf{Consistency}} &  \makecell[r]{\textbf{Affirmative}\\\textbf{Consistency}}  \\ 
  \midrule
\multicolumn{7}{c}{\cellcolor{light-gray} \emph{Heuristics}} \\
Majority & - & 47.75 & 1.35 & 51.50 & 16.48 & 8.71 \\ 
Question-Only  & 770M &52.32 & 11.80 & 48.15 & 24.42 & 24.02 \\ 
Edit-Type Only  & 770M &53.85 & 12.44 & 50.54 & 25.83 & 25.26  \\
Negation-Cue Only  & 770M &56.79 & 14.89 & 55.96 & 29.17 & 27.89\\ 
 
    \multicolumn{7}{c}{\cellcolor{light-gray} \emph{Fully Supervised}} \\

\textsc{BERT-Large} & 340M & 46.3 & 2.20 & 44.21 & 14.76 & 12.35 \\ 
\textsc{RoBERTa-Large} & 355M  & 54.08 & 13.64 & 51.64 & 26.53 & 27.18 \\ 
\textsc{DeBERTa-v2-XLarge} & 710M & 54.01 & 13.37 & 52.72 & 25.61 & 25.69  \\ 
\textsc{DeBERTa-v3-Large}  & 304M & 57.09 & 18.02 & 56.50 & 30.13 & 30.93  \\

\textsc{UnifiedQA-v2-Base}  & 220M & 57.94 & 17.49 & 54.62 & 30.39 & 32.98 \\ 
\textsc{UnifiedQA-v2-large}  & 770M & 66.72 & 30.20 & 63.98 & 43.68 & 46.45 \\ 
\textsc{UnifiedQA-v2-3B}  & 3B & \textbf{73.26} & \textbf{42.18} & \textbf{72.80} & \textbf{55.68} & \textbf{57.22} \\ 
    \multicolumn{7}{c}{\cellcolor{light-gray} \emph{Few-Shot}} \\

\textsc{UnifiedQA-v2-Base}  & 220M & 52.58	& 11.97	& 50.11	& 24.19	& 25.03 \\ 
\textsc{UnifiedQA-v2-Large} & 770M & 55.84 & 16.80	& 56.05	& 30.25 & 29.93							\\ 
\textsc{UnifiedQA-v2-3B} & 3B & 61.14 & 22.52 & 62.05	& 35.71& 35.41							 \\ 
\textsc{GPT-3} & 175B & 52.42 & 5.22 & 48.94 & 23.31 & 20.31 \\
\textsc{\faMagic InstructGPT} & N/A & 60.88 & 20.30 & 63.92 & 36.40 & 33.98 \\
\textsc{\faMagic InstructGPT + CoT} & N/A & \textbf{66.28} & \textbf{27.28} & \textbf{64.27} & \textbf{45.08} & \textbf{44.91} \\
\multicolumn{7}{c}{\cellcolor{light-gray} \emph{Zero-Shot}} \\
\textsc{UnifiedQA-v2-Base}  & 220M & 55.65& 16.20 & 52.47 & 29.23 & 30.83 \\ 
\textsc{UnifiedQA-v2-Large} & 770M & 61.74 & 23.07 & 61.16 & 37.14 & 37.14\\ 
\textsc{UnifiedQA-v2-3B} & 3B & 69.41 & 34.91 & 70.71 & 47.94 & 49.74\\ 
\textsc{UnifiedQA-v2-11B} & 11B & \textbf{73.11} & \textbf{40.02} & \textbf{75.48} & \textbf{53.72} & \textbf{54.12}\\ 
\textsc{FLAN-T5-XXL} & 11B & 67.53 & 31.61 & 67.43 & 45.45 & 47.86\\ 
\textsc{GPT-3} & 175B & 43.72 & 1.28 & 41.33 & 10.67 & 10.89 \\
\textsc{\faMagic InstructGPT} & N/A & 54.00 & 16.32 & 55.54 &29.87& 27.81 \\
\multicolumn{7}{c}{\cellcolor{light-gray} \emph{Human Performance}} \\
\textsc{Human}  & - & \textbf{91.94} & \textbf{81.58} & \textbf{93.65} & \textbf{86.49} & \textbf{88.22} \\
  \bottomrule
  \end{tabular}
}
\caption{Model performance on \datasetname. All heuristics are built on top of \textsc{UnifiedQA-Large}. \textbf{Boldface} indicates the best model on each metric for every training setup (\emph{Supervised}, \emph{Few-Shot}, \emph{Zero-Shot}). Supervised models are trained and evaluated across five random seeds using the full train and test sets. Due to the cost of OpenAI API, for few- and zero-shot models we report the average performance across five train-test splits. For more details and description of metrics see \S \ref{ref:baselines_and_metrics}. GPT-3 version: \texttt{davinci}; \faMagic InstructGPT version: \texttt{text-davinci-002}.}
\label{tab:ModelMainResults}

\end{table*}

\section{Results}

Model performance on \datasetname is given in Table~\ref{tab:ModelMainResults}. %
The best performing model is fully finetuned \uqathreeb with an accuracy of 73.26\% and overall consistency of 42.18\%, where the estimated human accuracy is 91.94\% and consistency 81.58\%. %
This gap shows that \datasetname questions are both answerable by humans, and challenging for state-of-the-art models. %

We create a contrastive dataset to be able to measure consistency as measuring models' ability to \emph{robustly} predict answers across small input perturbations can provide a more accurate view of linguistic capabilities~\cite{gardner-etal-2020-evaluating}. %
Here, there is a gap of $\sim$40\% in consistency between humans and the best model. %
Models are most robust to paraphrase edits: if a model answers a question correctly for the original passage, it is likely to be robust to changes in how that negation is expressed. %
We observe that the heuristic-based baselines exhibit low consistency, suggesting the consistency metric may be a more reliable measure than accuracy to evaluate models' ability to process negation. 
Thus, mainstream benchmarks should consider including consistency as a metric to more reliably measure progress on language understanding. 

Few- and zero-shot baselines do not match fully finetuned models' performance, but considerably improve over the majority baseline. %
For UnifiedQA-v2 in particular, this suggests that some reasoning about implications of negation is acquired during pretraining.
Surprisingly, UnifiedQA-v2 few-shot performance is worse than zero-shot. %
While this behavior has been reported for in-context learning with GPT-3 \cite{brown2020language, DBLP:conf/iclr/XieRL022}, we did not expect to observe this for a finetuned model.\footnote{A lower learning rate or less training steps do not help improve UnifiedQA-v2 few-shot performance.} %
UnifiedQA-v2-3B finetuned with a few examples is comparable to \faMagic InstructGPT (\texttt{text-davinci-002}; at least 175B parameters) with in-context learning. %
Chain-of-thought prompting (CoT) notably improves the performance of \faMagic InstructGPT, especially in terms of the most challenging metrics: scope and affirmative consistency. 
In the zero-shot setting, the 11B version of UnifiedQA-v2 performs the best, while the base version of only 220M parameters is comparable to \faMagic InstructGPT. 
UnifiedQA-v2-11B is also better than FLAN-T5-XXL (a 11B-parameter model as well). %
Given that UnifiedQA-v1 \cite{khashabi-etal-2020-unifiedqa} has been effective for tasks beyond QA \cite{DBLP:conf/nips/BraggCLB21, marasovic-etal-2022-shot-correct}, this result suggests that UnifiedQA models are strong but overlooked baselines in recent works on large-scale models. 

\section{Analysis}

While examining model errors, we find \uqalarge has a negative correlation with question length (Figure~\ref{ref:questionlength-modelperf} in Appendix~\ref{ref:model-analysis}). %
Humans can still reliably answer such long questions that are frequent in \datasetname. %
We also analyze the performance of \uqalarge across answer types, finding that: (i) the model performs best when the answer is ``No'', (ii) it almost never predicts ``Don't know'', and (iii) its performance on span extraction questions is in-between those two extremes (Figure~\ref{ref:answertype-modelperf} in Appendix~\ref{ref:model-analysis}). \uqathreeb exhibits similar behavior, with improved performance on questions which admit ``Don't know'' as an answer.

We analyze questions across the Wikipedia passages and the passages with edited scopes, with the focus on: (i) instances where the true answer does not change with the edited scope and the model should be stable, and (ii) instances where the true answer does change and the model should be sensitive to the edit. %
We find that when the fully-finetuned \uqathreeb  (the best-performing model) answers the question correctly for the Wikipedia passage, it only produces the answer correctly for 63.23\% of questions where the scope edit induces a different answer. In contrast, the model answers questions correctly for 91.03\% of the instances where the answer does not change with the scope edit.\footnote{Computed over the subset of questions which had high agreement for all four passages.} %
This suggests the model is not sensitive to changes of the scope of negated statements. 

We also analyze to what extent \uqathreeb distinguishes between negated statements and their affirmative counterparts. %
We examine model predictions for 1080 sample pairs where the answer changes when the negation is undone. %
For 43.52\% of these, the model changes its predictions. %
This suggests, in contrast to previous work~\cite{kassner-schutze-2020-negated, ettinger-2020-bert}, that models are sensitive to negated contexts to some extent.

\section{Related Work}
\label{sec:related_work}

In Aristotle's \emph{de Interpretatione}, all declarative statements are classified as either affirmations or negations used to affirm or contradict the occurrence of events~\cite{ackrill1975categories}. %
Negation is expressed through a variety of formulations~\cite{horn1989natural} and is prevalent in English corpora~\cite{hossain-etal-2020-analysis}.  %
Despite that, evidence from multiple tasks that require language understanding capabilities---such as NLI~\cite{naik-etal-2018-stress}, sentiment analysis~\cite{li-huang-2009-sentiment,zhu-etal-2014-empirical,barnes-etal-2019-sentiment}, paraphrase identification~\cite{kovatchev-etal-2019-qualitative}, machine translation~\cite{fancellu-webber-2015-translating-negation,hossain-etal-2020-non}, and QA~\cite{ribeiro-etal-2020-beyond,sen-saffari-2020-models}---identify negation as a challenging semantic phenomenon for models. \citet{hossain-etal-2022-analysis} analyze negation in 8 NLU datasets 
and conclude: ``new corpora accounting for negation are needed to solve NLU tasks when negation is present''. We expect \datasetname will help.


\paragraph{Negation Annotations} \citet{jimenez-zafra-etal-2020-corpora} overview datasets with negation as the main phenomenon and mention the following: BioScope \cite{DBLP:journals/bmcbi/VinczeSFMC08}, ProbBank Focus \cite{blanco-moldovan-2011-semantic}, ConanDoyle-neg \cite{morante-daelemans-2012-conandoyle}, SFU Review$_{\text{EN}}$ \cite{konstantinova-etal-2012-review}, NEG-DrugDDI \cite{bokharaeian-diaz-2013-nil}, NegDDI-Drug \cite{Bokharaeian2014ExploringNA}, and DT-Neg \cite{banjade-rus-2016-dt}. %
These datasets are small (<4K) and annotated with different schemes and guidelines as there is no established formalism for negation due to its complexity---the case when the QA format is useful \cite{DBLP:journals/corr/abs-1909-11291}. 
There are datasets focused on negation cue/scope/focus detection, or negated event recognition \cite{morante-blanco-2012-sem, reitan-etal-2015-negation, fancellu-etal-2017-detecting, he-etal-2017-neural,  li-lu-2018-learning, hossain-etal-2020-predicting}.
\citet{jimenez-zafra-etal-2020-corpora} assert that the lack of large datasets remains a major obstacle. 

\paragraph{Probing Negation} 
\citet{ettinger-2020-bert} introduces a 
dataset of 72 sentences for probing understanding of negation. %
\citet{kassner-schutze-2020-negated} analyze factual knowledge in the presence of negation. \change{Several works have recently constructed challenge sets that focus on negation for existing NLI datasets \cite{cooper1996using,dagan2006pascal,giampiccolo2007third}.} 
\citet{hartmann-etal-2021-multilingual} introduce a multilingual dataset for probing 
negation 
based on XNLI/MNLI \cite{conneau-etal-2018-xnli, williams-etal-2018-broad}. 
\citet{hossain-etal-2020-analysis} analyze negation in three existing NLI datasets and find they are unsuitable for studying how NLI models handle negation. 
They introduce a new benchmark of 4.5K instances based on 1.5K seed instances from the three NLI  datasets. \change{ \citet{geiger-etal-2020-neural} construct a dataset targeting the interaction between lexical entailment and negation, finding that models trained on general-purpose NLI datasets do not perform well, but finetuning with their dataset is sufficient to address this failure. %
In contrast to several of these works, we contribute training data and find that simply finetuning on these examples is not sufficient to address the challenges in \datasetname. %
See  Appendix \S\ref{ref:priorwork} for a detailed comparison.}

\paragraph{Improving  Negation Understanding}
 Efforts to improve models' negation abilities that can be studied on \datasetname are: unlikelihood training~\cite{hosseini-etal-2021-understanding}, NLI data~\cite{kim-etal-2019-probing}, commonsense knowledge~\cite{jiang-etal-2021-im}, multitasking~\cite{moore-barnes-2021-multi}, extra  MLM~\cite{khandelwal-sawant-2020-negbert,DBLP:journals/corr/abs-2205-04012}. %

\section{Conclusion}
Negation supports key properties of human linguistic systems such as the ability to distinguish between truth and falsity.  We present \datasetname, a QA dataset that contains \datasetsize examples to evaluate models' ability to reason about the implication of negated statements. We describe a procedure for contrastive dataset collection that results in challenging questions, present a detailed analysis of the dataset, and evaluate a suite of strong baselines in fully-finetuned, few-shot, and zero-shot settings. We evaluate models on both their accuracy and consistency, and find that this dataset is highly challenging---even the best-performing model is 18 points lower in accuracy than our human baseline, and about 40 points lower in consistency. %
We expect that \datasetname will facilitate NLU systems that can handle negation.

\section*{Limitations}
In this work, we contribute \datasetname, a dataset to facilitate the development of models that can process negation. Though \datasetname currently represents the largest NLU dataset that evaluates a model's ability to process the implications of negation statements, it is possible to construct a larger dataset, with more examples spanning different answer types. Further, \datasetname is an English dataset, and it would be interesting to extend our data collection procedures to build high-quality resources in non-English languages. Finally, while we attempt to extensively measure and control for artifacts in \datasetname, it is possible that the dataset has hidden artifacts that we did not study.

\section*{Acknowledgements}

The authors are grateful to Aakanksha Naik, Aditya Potukuchi, Rajat Kulshreshtha, Shruti Rijhwani, and Ashu Ravichander for feedback on the crowdsourcing task, and to Noah Smith and Jena Hwang for valuable discussions related to this project. The authors would also like to thank all members of the AllenNLP team and the three anonymous reviewers for their valuable feedback. 
\bibliography{anthology,custom}
\bibliographystyle{acl_natbib}

\clearpage 

\appendix

\definecolor{light-gray}{gray}{0.94}
\begin{table*}[!h]
         \centering
\resizebox{\textwidth}{!}{
  \begin{tabular}{lccccccp{5cm}c}
  \toprule
\textbf{Dataset}&  \makecell[c]{\textbf{Task}} & \makecell[c]{\textbf{Size}} & \makecell[c]{\textbf{Contrastive}\\ \textbf{Training}\\\textbf{ Data}}& \makecell[c]{\textbf{Passage}\\\textbf{/Premise}\\\textbf{/Prompt} \\ \textbf{Length}} & \makecell[c]{\textbf{Question}\\\textbf{/Hypothesis}\\ \textbf{Length}} & \makecell[c]{\textbf{\# Negation}\\ \textbf{Cues}} & \makecell[c]{\textbf{Data}\\ \textbf{Creation}}  &  \makecell[c]{\textbf{Answer }\\\textbf{ Exists}}  \\ 
  \midrule
\datasetname & RC  &\textbf{14,182} & \textbf{\cmark} & \textbf{132.50} & \textbf{24.44} & \textbf{219} &  Trained crowdworkers  (a) paraphrase negation, (b) change negation scope, (c) remove the negation, (d) ask questions about implications of negation, (e) provide answers, (f) verify answers  & \cmark \\   
\midrule
\newcite{hossain-etal-2020-analysis} & NLI & \makecell[c]{1500 (MNLI)\\1500 (SNLI)\\1500 (RTE)} & \xmark &  \makecell[c]{16.71 (MNLI)\\12.82 (SNLI)\\23.73 (RTE)}  & \makecell[c]{11.27 (MNLI)\\8.69 (SNLI)\\11.04 (RTE)} &  1 & Insert negation cue automatically  & \cmark\\   
\midrule
\newcite{geiger-etal-2020-neural} & NLI  & 2678   & \xmark & 9.27 &  9.27 & 1 & Fill template automatically using Wordnet~\cite{fellbaum1998semantic}  & \cmark \\   
\midrule
\newcite{hartmann-etal-2021-multilingual} & NLI  & 1960 & \xmark & 19.35  & 9.95 & 66 & Remove negation &  \cmark \\   \midrule
\newcite{ettinger-2020-bert} & Cloze task  & \makecell[c]{72 (NEG-136-SIMP)\\64 (NEG-136-NAT)} & \xmark & 5.5 / 7.5 & - & 1 & Psycholinguistic stimuli & \xmark \\   
  \bottomrule
  \end{tabular}
}
\caption{Comparison between \datasetname and prior datasets focusing on probing negation. We examine the English data in ~\citet{hartmann-etal-2021-multilingual}, the MNLI/SNLI/RTE splits in \citet{hossain-etal-2020-analysis}, NMoNLI~\cite{geiger-etal-2020-neural}, as well as the NEG-136-SIMP and NEG-136-NAT datasets~\cite{ettinger-2020-bert}. \datasetname is a reading comprehension dataset (RC), tasks in \citet{hartmann-etal-2021-multilingual} and \citet{hossain-etal-2020-analysis} are stress tests for existing general-purpose NLI datasets such as MNLI. NMoNLI is used both as a challenge (evaluation) set and to train models on a subset of the data. NEG-136-SIMP/NEG-136-NAT are datasets of cloze-style prompts. Passage/Premise/Prompt length and Question/Hypothesis length are described using the average number of words in the input. ``Answer exists'' describes whether a correct answer exists for the negated statement in the dataset, or if the evaluation relies on negated and affirmative statements requiring different predictions.}

\label{tab:comparison_priorwork}

\end{table*}

\section{Extended Comparison to Prior Negation Datasets}
\label{ref:priorwork}

In this section, we complement the discussion in \sect{sec:related_work} on how \datasetname differs from existing datasets focused on negation. %
A detailed comparison is given in Table \ref{tab:comparison_priorwork}. %

\begin{figure}[!h]
\centering
 \subfloat[Negation cue distribution in \citet{hartmann-etal-2021-multilingual}.]
 {
  \label{fig:synthetic_asr}
  \includegraphics[width=0.9\columnwidth]{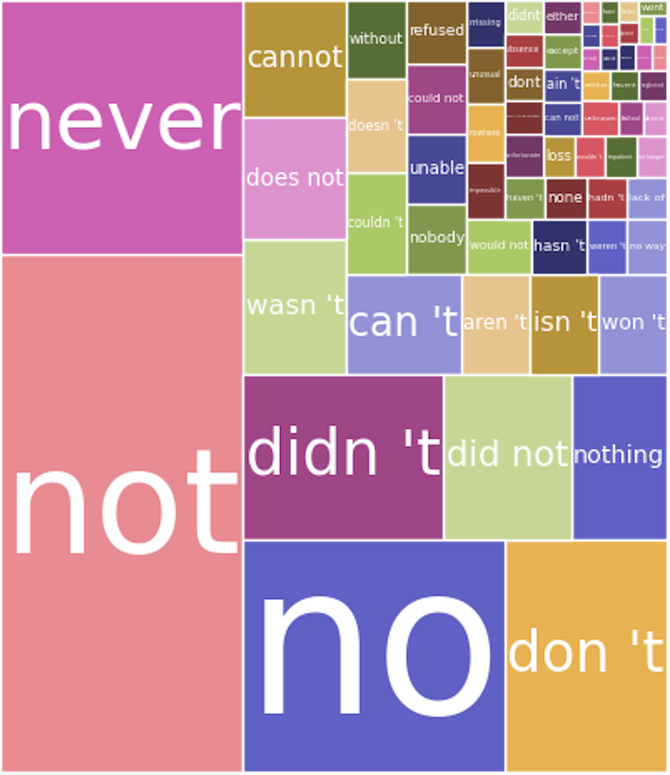}
  } \hspace{0.5cm}
  \subfloat[Negation cue distribution in \datasetname.] 
  {
  \label{fig:natural_asr}
  \includegraphics[width=0.9\columnwidth]{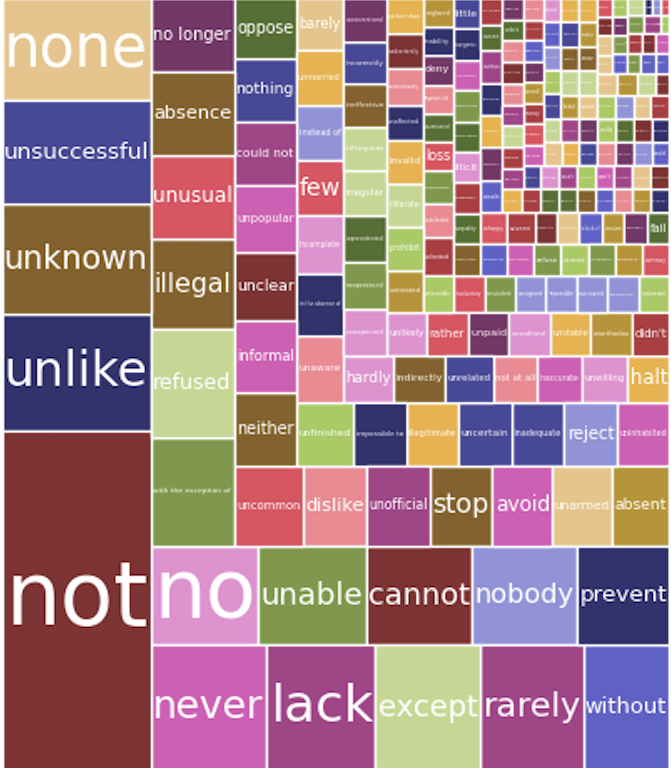}
  } 
\caption{Visualization of negation cues distributions.}
\label{fig:negcue_comp}
\end{figure}

Our goal with constructing \datasetname is to contribute a high-quality and systematic evaluation that will facilitate future models that can adequately process negation. %
To this end, we aim to construct a benchmark where artifacts are carefully mitigated~\cite{gardner-etal-2020-evaluating}, that is large enough to support robust evaluation, and that covers competencies any NLU system needs for adequate processing of negation. %
For example, the ability to recognize the implications of negated statements, distinguish them from their affirmative counterparts, and identify their scope. As such, main properties that \datasetname has compared to prior datasets focused on negation are: 
\begin{compactenum}
\item It is the first English reading-comprehension dataset that targets how models process negated statements in paragraphs~\cite{gardner2019question}. 
\item It features three types of contrastive inputs to test a model's sensitivity to the presence of negation, its exact scope, and the way it is phrased. As such, it is the first contrastive dataset for studying negation. 
\item It is substantially larger in size to facilitate robust evaluation.
\item It contains diverse forms of negation. Prior work constructing negation-based challenge sets for NLI models have largely constructed instances by using `not' as the only negation cue~\cite{hossain-etal-2020-analysis, naik-etal-2018-stress}. \newcite{hartmann-etal-2021-multilingual} extend this and include 66 English negation cues in their NLI challenge set. Our dataset consists of over 200 negation cues. Figures \ref{fig:synthetic_asr} and \ref{fig:natural_asr} illustrate the distribution of negation cues in the dataset by \citet{hartmann-etal-2021-multilingual} and \datasetname, respectively. \datasetname is less skewed toward a few negation cues such as ``not'', ``never'', ``no'', etc. %
\item All examples are manually constructed by well-trained crowdworkers rather than by using rules and templates.
\item It includes a rigorous validation procedure by several crowdworkers to mitigate examples being incorrect or ambiguous.  
\end{compactenum}


\section{Analysis of \datasetname}
\label{ref:data-analysis-appendix}

\paragraph{Commonsense Inferences}
We provide a categorization of the types of commonsense inferences required to answer \datasetname questions. %
These categories are presented in Table \ref{ref:qreasoning_types}. 

\paragraph{Edit Strategies}
We provide a set of edit strategies that were employed by crowdworkers to make paraphrase and scope edits. These edits are given in Table \ref{tab:EditTypes}.

\paragraph{Question/Passage Overlap} An issue with some NLU datasets is that simple heuristics based on lexical overlap are sufficient to achieve high performance~\cite{weissenborn-etal-2017-making,naik-etal-2018-stress}. %
We measure the lexical overlap between \datasetname questions and passages and find that is considerably lower than many prior QA datasets. %
Specifically, the average overlap between questions words and passage words is 0.52, which is lower compared to
SQuAD 1.0~\cite{rajpurkar-etal-2016-squad} (0.63), SQuAD 2.0~\cite{rajpurkar-etal-2018-know} (0.63), RACE~\cite{lai-etal-2017-race} (0.67), and Quoref~\cite{dasigi-etal-2019-quoref} (0.72).

\begin{figure}[tb]

    \includegraphics[width=\columnwidth]{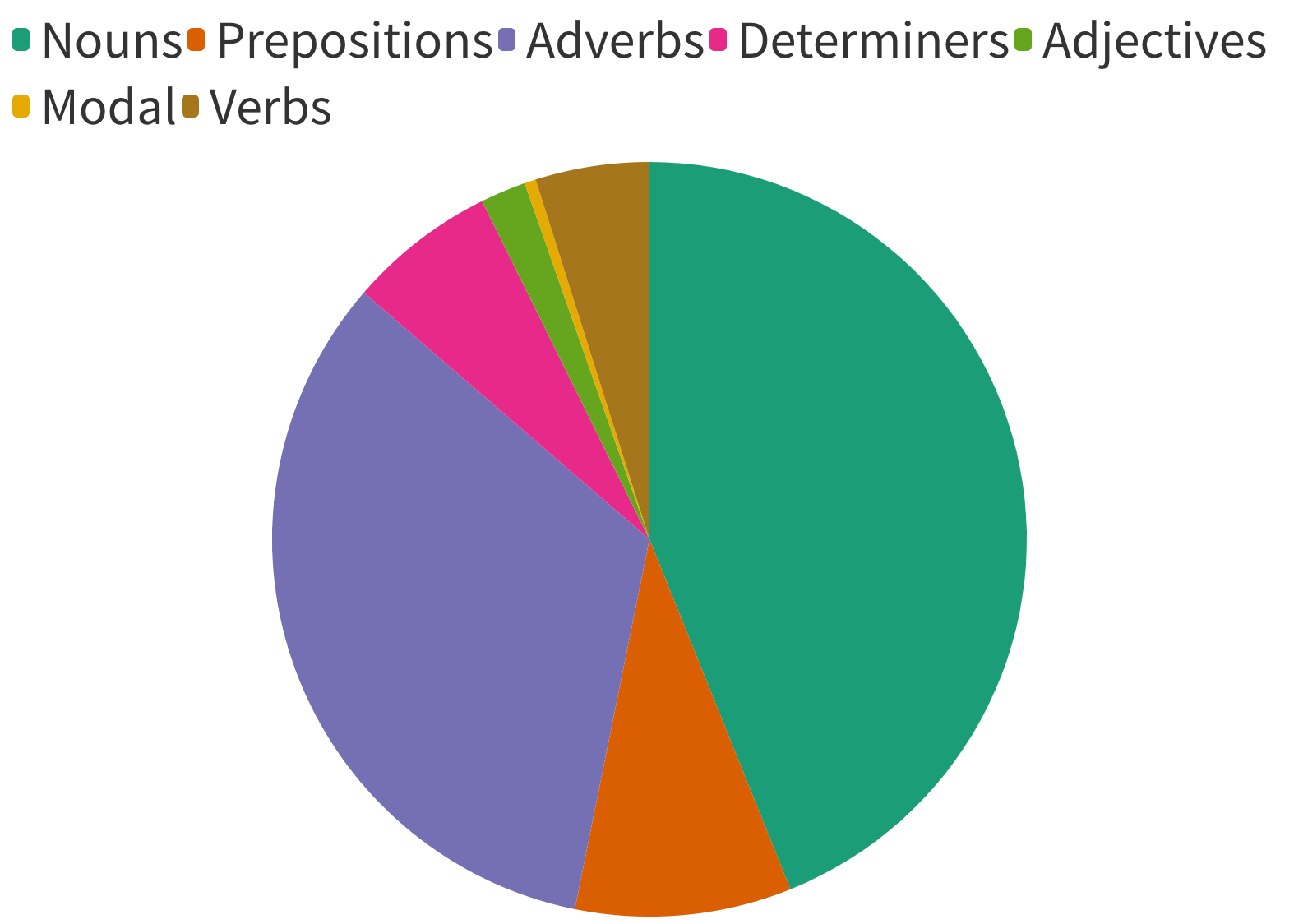}

  \caption{Distribution of grammatical categories of negation cues in \datasetname.  }
 \label{ref:negcue-grammar}
\end{figure}

\paragraph{Distribution of grammatical categories of negation cues}
We analyze the distribution over grammatical categories for single-word negation cues in \datasetname. We use the NLTK library~\cite{bird2009natural} to identify part-of-speech tags for these cues. These results are shown in Figure \ref{ref:negcue-grammar}.

\paragraph{Model sensitivity to edits} One potential issue with the dataset may be that models find it trivial to distinguish between edited passages and leverage this information to answer questions. %
To evaluate whether models can easily distinguish between the original passages and edited versions, we train BERT~\cite{devlin-etal-2019-bert} on the task of identifying whether a passage is sourced from Wikipedia or is an edited passage produced by a crowdworker. 
We expect it should be simple for these models to distinguish between the Wikipedia passages and the affirmative edits, as the model can simply rely on the presence or absence of a negation cue. %
We observe that as expected, models are somewhat able to distinguish the original Wikipedia passages from affirmative edits, but are largely unable to discriminate between the original passage and the paraphrase and scope edits (Table \ref{tab:EditClassifier}). 

\paragraph{Naturalness of edits}

New edits made by crowdworkers may contain unnatural sentences or linguistic constructs. To quantify this and to exclude the possibility that model performance degrades only due to the unnaturalness of the edits, we compare the perplexity assigned by the OpenAI-GPT language model~\cite{radford2018improving} to the edited passages and the original Wikipedia passages, finding that they are largely similar (Table~\ref{tab:EditUnNatural}).

\begin{table}[!t]
\resizebox{\columnwidth}{!}{
\begin{tabular}{l@{\hskip 0.1in}llll} \toprule
Model        &  All & Original-Pa. & Original-Sc. & Original-Aff. \\ \midrule
Majority     &  25.65\%   &     50\%                & 51.86\%               &       50.68\%               \\ \midrule
BERT    &    34.4\% &     53.95\%                &  55.27\%              &   63.25\%                   \\\bottomrule   
\end{tabular}
}
\caption{Performance of models trained to distinguish Wikipedia text from edits made by crowdworkers. We used Bert-base, averaged over three random seeds. }
\label{tab:EditClassifier}
\end{table}
\begin{table}[tb]
\resizebox{\columnwidth}{!}{
\begin{tabular}{lcccc} \toprule
\textbf{Split}        & \textbf{Original} & \textbf{Paraphrase} & \textbf{Scope} & \textbf{Affirmative} \\ \midrule
Train    &  77.29   &     76.75                & 77.64              &       78.27              \\ \midrule
Dev    &  71.23   &     70.85               & 72.79              &       71.60              \\ \midrule
Test    &  74.38  &     74.88               & 75.63               &       76.05               \\ \bottomrule
\end{tabular}
}
\caption{Average perplexities of original and (paraphrase, scope, affirmative) edited passages calculated with  OpenAI-GPT~\cite{radford2018improving}.}
\label{tab:EditUnNatural}
\end{table}

\paragraph{Consistency Groups}
We provide data statistics on the instances that are used to compute consistency metrics on the dataset. There are 5,608 instances  in the dataset that are included in consistency groups, and thus there are 1,402 ``groups'' to compute question-level consistency. and each edit-level consistency metric.

\section{Model Training Details}
\label{sec:appendix}

All models we evaluate on \datasetname are pretrained transformer-based language models. %
We test them in three training settings: (ii) finetuned on the entire training data (\sect{sec:appendix_supervised_models}), (ii) finetuned on a few examples (few-shot; \sect{sec:appendix_few_zero_shot_details}), and (iii) without training (zero-shot; \sect{sec:appendix_few_zero_shot_details}).

\subsection{Fully Finetuned}
\label{sec:appendix_supervised_models}

We train all fully-finetuned model with five seeds and report the average performance across them. %
For every seed, we evaluate the model with the best validation accuracy on the entire test set. %

\paragraph{BERT~\cite{devlin-etal-2019-bert}} BERT is pretrained with masked language modeling (MLM) and a next-sentence prediction objective. %
Since a majority of the questions have Yes/No/Don't know as the answer, we finetune BERT and other BERT-like models (see below) in a multi-class classification setting. %
We train all BERT-like models in this fashion. %
In our experiments, we BERT-Large.  %
We train with a learning rate of 1e-5 for 10 epochs.

\paragraph{RoBERTa~\cite{liu2019roberta}} RoBERTa is a more robustly pretrained version of BERT. %
In our experiments, we use RoBERTa-Large.

\paragraph{DeBERTa~\cite{he2020deberta, he2021debertav3}} DeBERTa has a disentangled attention mechanism and it is pretrained with a version of MLM objective that uses the content and position of the context words. %
In our experiments, we use DeBERTa-v2-XLarge and DeBERTa-v3-Large.  %

\paragraph{UnifiedQA~\cite{khashabi-etal-2020-unifiedqa, khashabi2022unifiedqa}} UnifiedQA is built on top of the T5 architecture~\cite{raffel2020exploring} by further training it on 20 QA datasets. %
We use UnifiedQA-v2 and finetune it with a learning rate of 5e-5 for 5 epochs. %
In the fully-finetuned setting, we study Base, Large, and 3B versions of UnifiedQA-v2.

\begin{table*}[t]
\centering
\resizebox{0.8\textwidth}{!}{
\begin{tabular} {rrrrrr}
  \toprule
\makecell[r]{\textbf{Sampling}\\\textbf{Strategy}} & \textbf{Accuracy} & \textbf{Consistency} & \makecell[r]{\textbf{Paraphrase}\\ \textbf{Consistency}}  &  \makecell[r]{\textbf{Scope}\\\textbf{Consistency}} &  \makecell[r]{\textbf{Affirmative}\\\textbf{Consistency}}  \\ 
\midrule
1                 & \textbf{52.81}    & \textbf{6.62}        & \textbf{49.83}                           & \textbf{21.95}                      & \textbf{21.95}                            \\
2                 & 51.42    & 5.57        & 44.95                           & 21.25                      & 26.48                            \\
3                 & 50.31    & 4.88        & 40.07                           & 18.12                      & 25.78       \\
\bottomrule
\end{tabular}
}
\caption{Few-shot results of GPT-3 (\texttt{davinci}) on one split of the test data (1/5 of the entire test set, $\sim$1440 examples) using different strategies for sampling few shots. See \sect{sec:appendix_few_zero_shot_details} for descriptions of the sampling strategies.}
\label{tab:appendix_few_examples_sampling_strategy}
\end{table*}
\begin{table*}[t]
\centering
\resizebox{0.8\textwidth}{!}{
\begin{tabular} {rrrrrr}
  \toprule
\textbf{Max Seq Len} & \textbf{Accuracy} & \textbf{Consistency} & \makecell[r]{\textbf{Paraphrase}\\ \textbf{Consistency}}  &  \makecell[r]{\textbf{Scope}\\\textbf{Consistency}} &  \makecell[r]{\textbf{Affirmative}\\\textbf{Consistency}}  \\ 
\midrule
2045 & 60.88        & 20.30                        & 63.92 & 36.40                                  & 33.98                                                            \\
4000 & 59.70              & 20.42                  & 62.94                                                           & 36.04                                               & 34.38                              \\
\bottomrule
\end{tabular}
}
\caption{``InstructGPT'' (\texttt{text-davinci-002}) performance on one split of the test data (1/5 of the entire test set, $\sim$1440 examples) with more and less examples in the context. The average number of shots that fit in 2045 tokens (\texttt{davinci} max.\ input length) is 8--9, and 17-18 if the context is 4000 tokens (\texttt{text-davinci-002} max.\ input length).}
\label{tab:appendix_seq_len}
\end{table*}
\begin{figure*}[t]
\includegraphics[width=\textwidth]{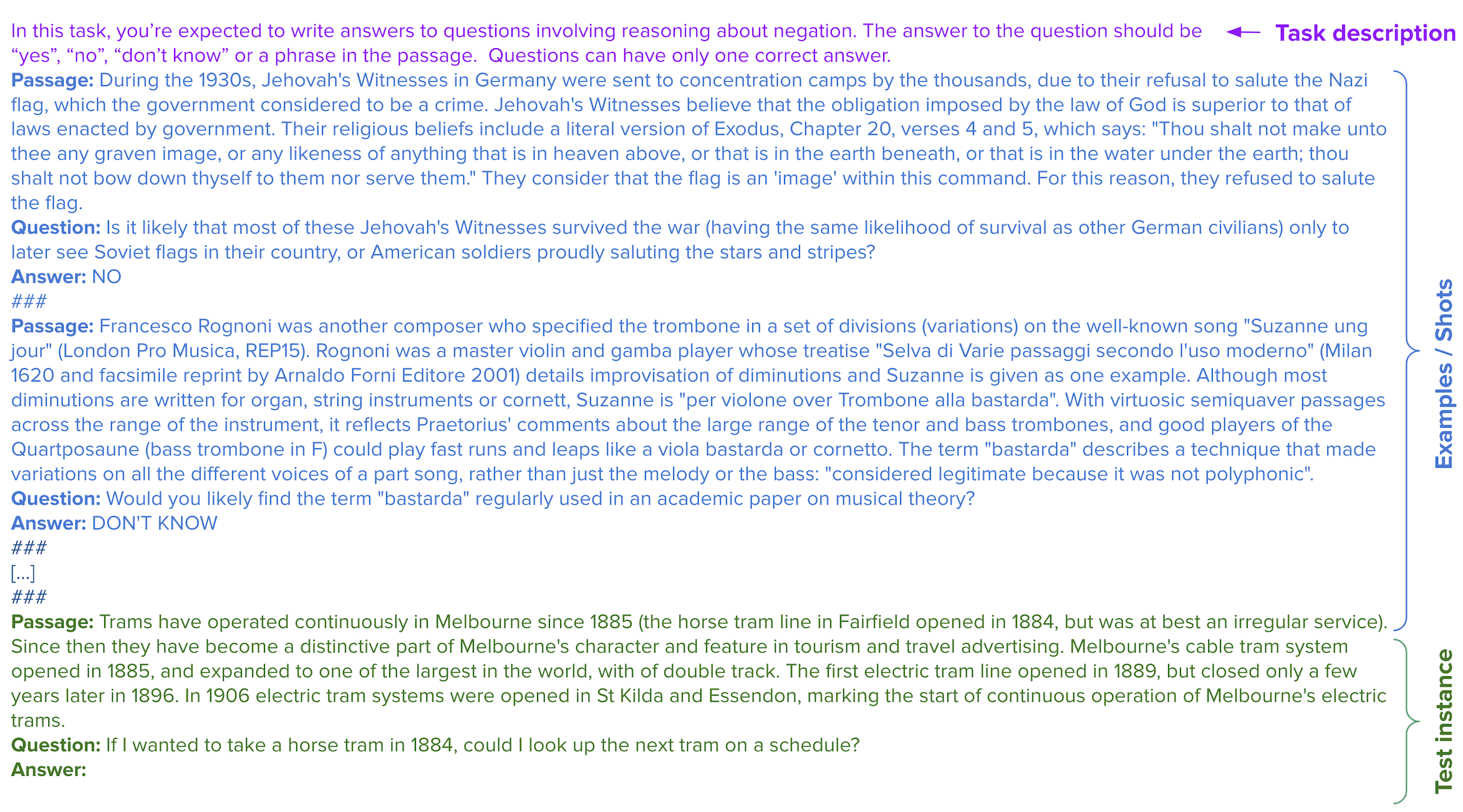}
\caption{A prompt used to get generations from GPT-3 (\texttt{davinci}) and ``InstructGPT'' (\texttt{text-davinci-002}). We designed the task description following \citet{DBLP:journals/corr/abs-2204-07705}. The zero-shot prompt is the same except that there are no examples.}
 \label{fig:prompt_davinci}
\end{figure*}

\begin{figure*}[t]
\includegraphics[width=\textwidth]{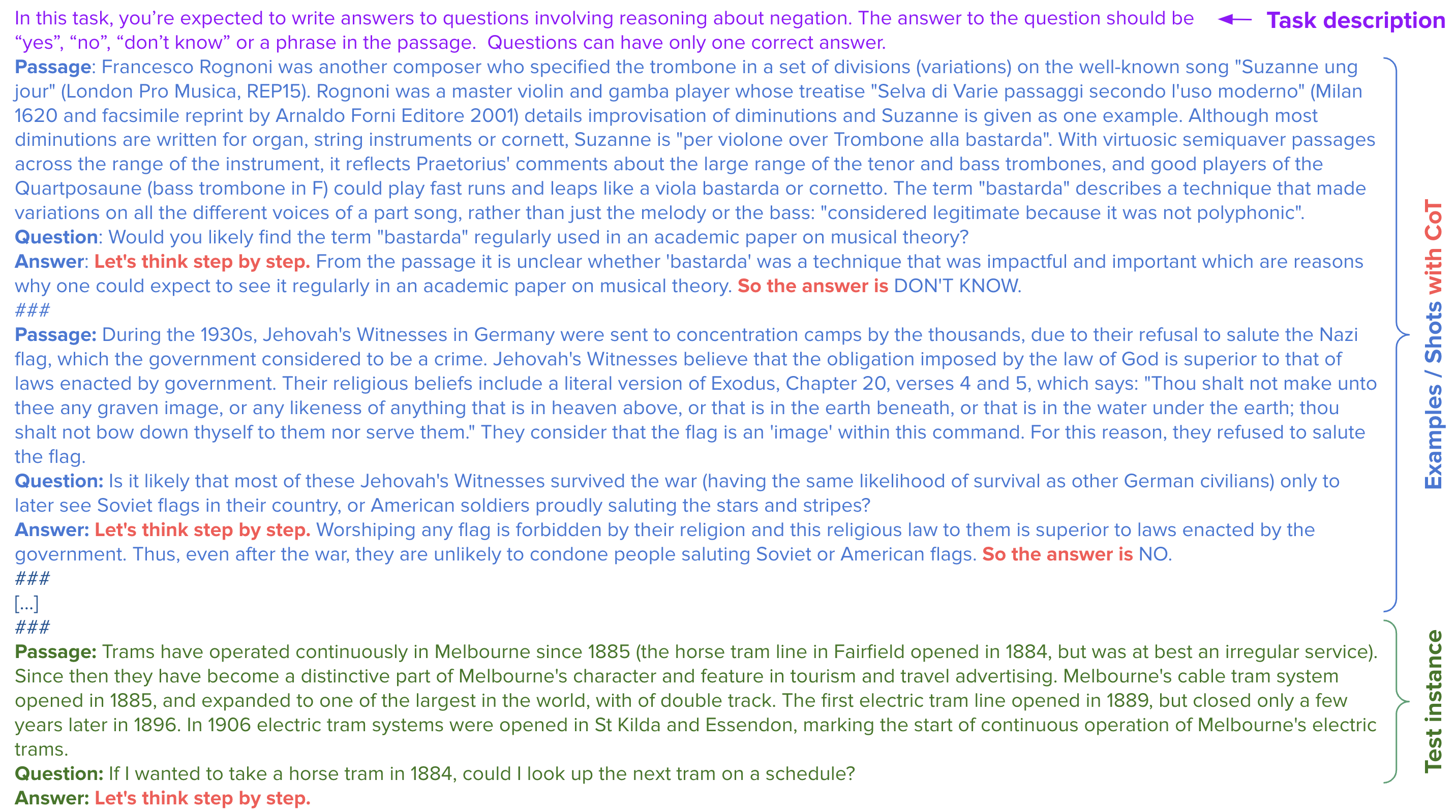}
\caption{A \emph{chain-of-thought} prompt (includes ``Let's think step by step. \{\texttt{explanation}\}. So the answer is'') used to get generations from ``InstructGPT'' (\texttt{text-davinci-002}). We designed the task description following \citet{DBLP:journals/corr/abs-2204-07705}.}
 \label{fig:prompt_cot}
\end{figure*}

\subsection{Few-shot and Zero-Shot}
\label{sec:appendix_few_zero_shot_details}

Unlike fully-finetuned models, we evaluate few- and zero-shot models on 5 train-test splits due to the cost of the OpenAI API. %
Evaluation on multiple disjoint splits of test data (that in union form the entire test set) with different choices of shots allows us to consider in our evaluation the sensitivity of few-shot learning to the choice of few examples. %
If the cost was not a concern, we would use five sets of few training examples and the \emph{entire} test set.

\paragraph{GPT-3~\cite[\texttt{davinci};][]{brown2020language}} This is the original GPT-3 model trained using only the standard LM objective. %
Its maximum input sequence length is $\sim$2K tokens which allows to fit on average 8--9 \datasetname training examples. Thus, we use this number of shots for few-shot experiments. %
To benchmark GPT models, we use the OpenAI API (in October 2022). %
We show one prompt for few-shot GPT models in Fig.\ \ref{fig:prompt_davinci}.

\paragraph{\faMagic InstructGPT~\cite[\texttt{text-davinci-002};][]{DBLP:journals/corr/abs-2203-02155}} This GPT variant does not come with a corresponding paper and little is known about it. %
It has recently been confirmed that it is an Instruct model, but unlike the original InstructGPT$_\textit{orig}$ \cite[\texttt{text-davinci-001};][]{DBLP:journals/corr/abs-2203-02155} it is not derived from GPT-3 (\texttt{davinci}).\footnote{\url{https://twitter.com/janleike/status/1584681562318458880}} %
InstructGPT$_\textit{orig}$ has been trained on the data that includes ``prompts submitted to earlier versions of the InstructGPT models on the OpenAI API Playground''.  
InstructGPT$_\textit{orig}$ is finetuned with reinforcement learning from human feedback \cite{DBLP:journals/corr/abs-2009-01325}. %
\texttt{text-davinci-002} has two times longer maximum input sequence length than \texttt{davinci} suggesting that the overall model size is notably larger too. %
This also means we can fit more examples in the context, but we do not find that to improve \texttt{text-davinci-002}'s performance; see Table \ref{tab:appendix_seq_len}. %
It has been reported on social media that \texttt{text-davinci-002} has notably stronger performance than \texttt{text-davinci-001}, but where do these improvements come from is publicly unknown.\footnote{\url{https://twitter.com/ben_bogin/status/1532022804886978568}} %

\paragraph{Chain-of-Thoughts (CoT) prompting \cite{DBLP:journals/corr/abs-2201-11903}} This type of prompting makes the model explain its prediction before providing it. %
When it was introduced, CoT prompting demonstrated benefits for math and commonsense reasoning. %
Since then, \citet{DBLP:journals/corr/abs-2210-09261} report that CoT prompting gives substantial improvements for a hard subset of the BIG-Bench tasks \cite{DBLP:journals/corr/abs-2206-04615}.\footnote{Another work shows limitations of prompting with explanations \cite{Ye-Durrett:2022:Fewshot}.} %
This makes it a promising prompt for our proposed task of reasoning about implications of negation. The suggested way to conduct CoT prompting (and how we use it in this paper)  is as follows:  
\begin{compactitem}
\item \textbf{Input:} \texttt{\{task\_description\}} \texttt{\{task\_examples\}} \texttt{\{test\_instance\}} \emph{Answer: Let's think step by step.}
\item \textbf{Output:} \texttt{\{explanation\}} \emph{So the answer is } \texttt{\{answer\}}
\end{compactitem}
One of the authors wrote explanations for all shots in each split (45 explanations in total) in few hours. %
In Figure \ref{fig:prompt_cot}, we show an example of a CoT prompt we use for ``InstructGPT'' (\texttt{text-davinci-002}).

\paragraph{FLAN-T5~\cite{Chung2022ScalingIL}} FLAN-T5 is a T5 variant that is further trained with instruction finetuning that includes CoT prompting, on over 1.8K tasks. %
We prompt FLAN-T5 in the zero-shot setting by constructing each test instance as follows:
\begin{compactitem}
\item \textbf{Input:} \emph{Passage:} \{\texttt{passage}\}\textbackslash n\emph{Question:} \{\texttt{question}\}\textbackslash n\emph{Give the rationale before answering.} 
\item \textbf{Output:} \texttt{\{explanation\}} \emph{So the answer is } \texttt{\{answer\}}.
\end{compactitem}
This output form is the most common, but the model sometimes generates ``(final) answer is'', ``(final) answer:'', etc., instead of ``So the answer is''. 

\paragraph{UnifiedQA-v2~\cite{khashabi2022unifiedqa}} We also evaluate UnifiedQA-v2 in a few- and zero-shot settings in addition to fully training it. We construct instances  following how they are constructed for training UnifiedQA-v2:
\begin{compactitem}
\item \textbf{Input:} \{\texttt{passage}\}\textbackslash n\{\texttt{question}\}
\item \textbf{Output:} \{\texttt{answer}\}
\end{compactitem}
We normalize and lowercase passages, questions, and answers. %
We manually choose hyperparameters following \citet{DBLP:conf/nips/BraggCLB21} and keep them fixed.

\paragraph{Which few examples to select?}  \datasetname's  unique structure raises the question of which 8--9 examples to use for few-shot learning: 
\begin{compactenum}
\item Randomly selected, 
\item Random without affirmative paragraphs to include more paragraphs with negation cues, 
\item Two groups of two questions and corresponding 4 paragraphs (original and three edited), 
\item Three groups of two questions and corresponding 3 paragraphs (original, scope- and paraphrase-edited; no affirmative).
\end{compactenum}
We hypothesize that the last two options could be beneficial for consistency of few-shot models. %
We prompt \texttt{davinci} with 1st and 3rd options, and depending which is better we evaluate 2nd or 4th (i.e., the better option without affirmative paragraphs). %
Contrary to our expectations, we find that the 1st option works better than 3rd, as well as better than the 2nd option; see Table \ref{tab:appendix_few_examples_sampling_strategy}. %
Therefore, for each training split, we sample 9 paragraph-question pairs randomly (sometimes only 8 fit in the context) and use these samples for all few-shot experiments.

\section{Model analysis}
\label{ref:model-analysis}

\paragraph{Model performance stratified by passage type} 
In Table \ref{tab:accuracy_wrt_edit_type}, we report the accuracy of model predictions corresponding to the type of passage: i.e whether the question was asked on the original Wikipedia passage, its paraphrase edit, its scope edit or the affirmative edit. \begin{figure}[tb]
    \includegraphics[width=\columnwidth]{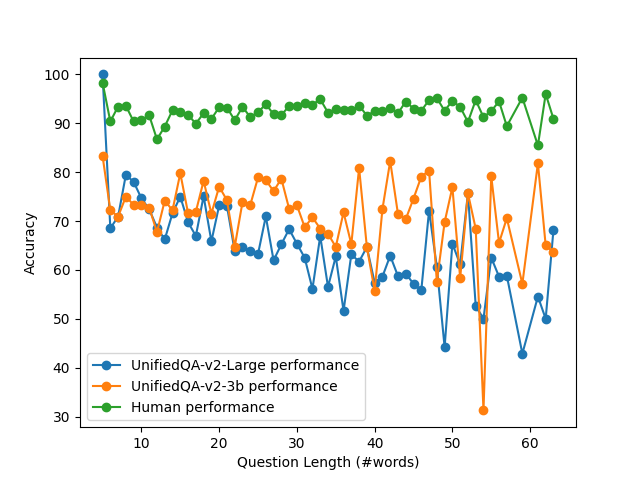}
  \caption{\textsc{UnifiedQA-v2-Large} and \textsc{UnifiedQA-v2-3b} performance stratified by the length of the question.}
 \label{ref:questionlength-modelperf}
\end{figure}
When we compare those results with those in Table \ref{tab:ModelMainResults}, we observe that fully-finetuned UnifiedQA-v2 shows largely similar QA performance in terms of accuracy on these different passage types, despite having very different consistency scores with the original passage. %
In contrast, GPT-3 and \faMagic Instruct-GPT in the few-shot setting perform better on the original Wikipedia passages and their paraphrased versions than on the scope and affirmative edits.

\begin{table}[t]
\resizebox{\columnwidth}{!}{
\begin{tabular}{lcccc} \toprule
\textbf{Model}              & \textbf{Original} & \textbf{Paraphrase} & \textbf{Scope} & \textbf{Affirmative} \\ \midrule
UnifiedQA-V2-3B    & 75.53 & 74.23 & 69.42 & 71.43   \\
UnifiedQA-V2-Large & 68.35 & 68.13 & 63.25 & 67.22      \\
 \textsc{GPT-3}   &  57.67 & 59.79 & 51.32 & 43.91              \\
  \textsc{\faMagic InstructGPT} & 67.99 & 70.63 & 53.37 & 51.84             \\
\bottomrule        
\end{tabular}
}
\caption{Accuracy of \textsc{UnifiedQA-V2}, \textsc{GPT-3}, and \textsc{\faMagic Instruct-GPT} stratified by the type of passage.}
\label{tab:accuracy_wrt_edit_type}
\end{table}

\begin{figure}[tb]

    \includegraphics[width=\columnwidth]{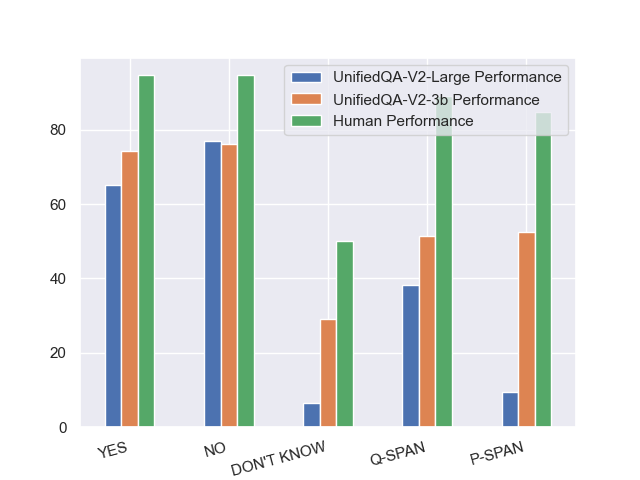}

  \caption{Model accuracy for \textsc{UnifiedQA-v2-Large} and \textsc{UnifiedQA-v2-3b} based on answer type.}
 \label{ref:answertype-modelperf}
\end{figure}

\paragraph{Model performance by question length}
In Figure \ref{ref:questionlength-modelperf}, we show model performance stratified by question length. %
We observe that longer questions are more difficult for fully-finetuned \textsc{UnifiedQA-v2-Large} but \textsc{UnifiedQA-v2-3B} appears to exhibit similar QA performance on some of these long questions.

\paragraph{Model performance by answer type}
In Figure \ref{ref:answertype-modelperf}, we show results of fully-finetuned model performance stratified by answer type (Figure \ref{ref:answertype-modelperf}). %

\paragraph{Variance in model performance}
We report the standard deviation of UnifiedQA-V2 models computed over the results from five \emph{seeds}, as well as the standard deviation of GPT-3 and \faMagic Instruct-GPT in few-shot and zero-shot settings computed over five \emph{splits}. These are shown in Table \ref{tab:ModelMainResultsVariance}.

\paragraph{Novelty of negation cues}

We compare the performance of fully-finetuned UnifiedQA-v2 Large/3B on Wikipedia passages where the negation cue has occurred in the training data, with the performance for novel negation cues. %
We find that model accuracy for UnifiedQA-V2-Large is 68.03 when the negation cue is unseen (has not been the cue in the negated statement that crowdworkers construct questions around in the training data), and 70.45 when it has appeared before in the training data. %
Similarly, UnifiedQA-V2-3B's accuracy is 74.38 and 73.73 for unseen and seen cues respectively. %
This suggests that the novelty of the negation cue is not a major factor of difficulty for UnifiedQA-v2 once it has been finetuned on the entire training data.

\definecolor{light-gray}{gray}{0.94}
\begin{table*}[!h]
         \centering
\small
\resizebox{\textwidth}{!}{
  \begin{tabular}
  {lcrrrrr}
  \toprule
\textbf{Model} & \textbf{\# Param} & \textbf{Accuracy} & \textbf{Consistency} & \makecell[r]{\textbf{Paraphrase}\\ \textbf{Consistency}}  &  \makecell[r]{\textbf{Scope}\\\textbf{Consistency}} &  \makecell[r]{\textbf{Affirmative}\\\textbf{Consistency}}  \\ 
  \midrule

    \multicolumn{7}{c}{\cellcolor{light-gray} \emph{Fully Finetuned}} \\

\textsc{UnifiedQA-v2-Base}  & 220M & 57.94$_{\pm 0.25}$ & 17.49$_{\pm 0.47}$ & 54.62$_{\pm 0.47}$ & 30.39$_{\pm 0.49}$ & 32.98$_{\pm 0.48}$ \\ 
\textsc{UnifiedQA-v2-large}  & 770M & 66.72$_{\pm 0.13}$ & 30.20$_{\pm 0.10}$ & 63.98$_{\pm 0.31}$ & 43.68$_{\pm 0.25}$ & 46.45$_{\pm 0.38}$ \\ 
\textsc{UnifiedQA-v2-3B}  & 3B & \textbf{73.26}$_{\pm 0.46}$ & \textbf{42.18}$_{\pm 0.72}$ & \textbf{72.80}$_{\pm 0.68}$ & \textbf{55.68}$_{\pm 0.58}$ & \textbf{57.22}$_{\pm 0.77}$ \\ 
    \multicolumn{7}{c}{\cellcolor{light-gray} \emph{Few-Shot}} \\
					

\textsc{UnifiedQA-v2-Base}  & 220M & 52.58$_{\pm 1.57}$	& 11.97$_{\pm 1.57}$	& 50.11$_{\pm3.32}$	& 24.19$_{\pm2.83}$	& 25.03$_{\pm 3.81}$ \\ 
\textsc{UnifiedQA-v2-Large} & 770M & 55.84$_{\pm 2.04}$	 & 16.80$_{\pm 2.28}$	& 56.05$_{\pm 2.96}$	& 30.25$_{\pm 2.01}$ & 29.93$_{\pm 3.14}$							\\ 
\textsc{UnifiedQA-v2-3B} & 3B & 61.14$_{\pm 3.45}$ & 22.52$_{\pm 5.2}$ & 62.05$_{\pm 2.82}$	& 35.71$_{\pm 3.66}$& 35.41$_{\pm 5.46}$							 \\ 
\textsc{GPT-3}$^*$ & 175B & 52.42$_{\pm 2.04}$ & 5.22$_{\pm 2.48}$ & 48.94$_{\pm 1.11}$ & 23.31$_{\pm 3.24}$ & 20.31$_{\pm 5.35}$ \\
\textsc{\faMagic InstructGPT$^{**}$} & N/A & 60.88$_{\pm 1.44}$ & 20.30$_{\pm 1.38}$ & 63.92$_{\pm 1.48}$ & 36.40$_{\pm 3.10}$ & 33.98$_{\pm 1.53}$ \\
\textsc{\faMagic InstructGPT$^{**}$+ CoT} & N/A & \textbf{66.28}$_{\pm 2.49}$ & \textbf{27.28}$_{\pm 3.85}$ & \textbf{64.27}$_{\pm 3.36}$ & \textbf{45.08}$_{\pm 2.82}$ & \textbf{44.91}$_{\pm 3.05}$ \\
\multicolumn{7}{c}{\cellcolor{light-gray} \emph{Zero-Shot}} \\
\textsc{UnifiedQA-v2-Base}  & 220M & 55.65$_{\pm 1.44}$& 16.20$_{\pm 1.74}$ & 52.47$_{\pm 2.366}$ & 29.23$_{\pm 1.27}$ & 30.83$_{\pm 1.95}$ \\ 
\textsc{UnifiedQA-v2-Large} & 770M & 61.74$_{\pm 0.8}$ & 23.07$_{\pm 2.39}$ & 61.16$_{\pm 2.58}$ & 37.14$_{\pm 1.3}$ & 37.14$_{\pm 2.93}$\\ 
\textsc{UnifiedQA-v2-3B} & 3B & 69.41$_{\pm 0.99}$ & 34.91$_{\pm 1.81}$ & 70.71$_{\pm 1.87}$ & 47.94$_{\pm 2.39}$ & 49.74$_{\pm 2.22}$\\ 
\textsc{UnifiedQA-v2-11B} & 11B & \textbf{73.11}$_{\pm 1.74}$ & \textbf{40.02}$_{\pm 2.84}$ & \textbf{75.48}$_{\pm 2.98}$ & \textbf{53.72}$_{\pm 2.32}$ & \textbf{54.12}$_{\pm 3.84}$\\ 

\textsc{FLAN-T5-XXL} & 11B & 67.53$_{\pm 1.25}$ & 31.61$_{\pm 3.37}$ & 67.43$_{\pm 2.54}$ & 45.45$_{\pm 3.27}$ & 47.86$_{\pm 2.58}$\\ 

\textsc{GPT-3}$^*$ & 175B & 43.72$_{\pm 0.86}$ & 1.28$_{\pm 0.35}$ & 41.33$_{\pm 2.60}$ & 10.67$_{\pm 1.90}$ & 10.89$_{\pm 1.282}$ \\
\textsc{\faMagic InstructGPT$^{**}$} & N/A & 54.00$_{\pm 2.05}$ & 16.32$_{\pm 2.95}$ & 55.54$_{\pm 2.56}$ &29.87$_{\pm 3.54}$& 27.81 $_{\pm 2.44}$\\

  \bottomrule
  \end{tabular}
}
\caption{Model performance on \datasetname with standard deviation. \textbf{Boldface} indicates the best model on each metric for every training setup (\emph{Supervised}, \emph{Few-Shot}, \emph{Zero-Shot}). Supervised models are trained and evaluated across five random seeds using the full train and test sets. Due to the cost of OpenAI API, for few- and zero-shot models we report the average performance across five train-test splits. For more details and description of metrics see \S \ref{ref:baselines_and_metrics}. GPT-3 version: \texttt{davinci}; \faMagic InstructGPT version: \texttt{text-davinci-002}.}
\label{tab:ModelMainResultsVariance}

\end{table*}

\begin{table*}[!h]
\centering
\small
\begin{tabular}{p{0.08\textwidth}p{0.35\textwidth}p{0.25\textwidth}p{0.05\textwidth}p{0.15\textwidth}} \toprule
\textbf{Reasoning Type}     & \textbf{Passage Snippet} & \textbf{Question} & \textbf{Answer} & \textbf{Explanation} \\ 
\toprule
\emph{Precondition}   (12\%)       & \textcolor{blue}{At first reluctantly but then with increasing vigour, Galen promoted Hippocratic teaching, including venesection and bloodletting, then  \textbf{unknown} in Rome }[...]  &   Would doctors in Rome regularly have performed venesection?    & NO & People can’t do a complicated procedure that they don’t know. \\ 
\midrule 
\emph{Social Norms}   (10\%)      &  \textcolor{blue}{On October 8, 1883, the US patent office ruled that Edison's patent was based on the work of William E. Sawyer and was, therefore, \textbf{invalid} .} Litigation continued for nearly six years. In 1885, Latimer switched camps and started working with Edison.         &  From the information given in the passage, would you say that coincidence is the most charitable explanation for what was essentially the same innovation, in much the same way that Newton and Leibniz seemingly discovered calculus independently, without knowing of the other's progress?       &  YES   & Plagarism is frowned upon in society, more so than accidentally reaching the same conclusions as someone else.      \\ 
\midrule
\emph{Psychology}  (9\%)     &    [...] Disraeli later romanticised his origins, claiming his father's family was of grand Iberian and Venetian descent; in fact Isaac's family was of no great distinction [...] \textcolor{blue}{ Historians differ on Disraeli's motives for rewriting his family history: [...] Sarah Bradford believes "his  \textbf{dislike}  of the commonplace would not allow him to accept the facts of his birth as being as middle-class and undramatic as they really were". }        &   Would Disraeli have been flattered by a biography that explored his middle class upbringing, according to Bradford?      & NO   &  A person such as Disraeli who wants to project a grandiose image of themselves is likely to be unhappy when people discuss mundane aspects about his upbringing.     \\ 
\midrule
\multirow{2}{2cm}{\emph{Cause and Effect} (7\%) }         &     
Oil produced from palm fruit is called `red palm oil' or just `palm oil'... \textcolor{blue}{ In its \textbf{unprocessed}  state, red palm oil has an intense deep red color because of its abundant carotene content.} [...] &   Would a consumer who was primarily interested in the eye-health benefits of carotenes and lycopene want to shop for palm oils by their color, rather than listening to marketing slogans such as "extra virgin" or "minimally processed"?   & YES & A high carotene content causes a deep red color, so a person searching for things with high carotene content can look at their color.  \\
\midrule
\multirow{2}{2cm}{\emph{Mutual Exclusivity} (5\%) } &   [...] \textcolor{blue}{The waterway system covered much of the country, and by the 1980s Finland had extended roadways and railroads to areas \textbf{not}  served by waterways, effectively opening up all of the country's forest reserves to commercial use. }          &    Would a person in 1990 taking a nap near a river in Finland be likely to be woken up by a train horn?     & NO  &  It is likely that the government prioritized building roads and railways in places not near waterways      \\ 
\midrule
\emph{Synecdoche}   (2\%)       &     \textcolor{blue}{ Al-Libi told the interrogators details about Richard Reid, a British citizen who had joined al-Qaeda and trained to carry out a suicide bombing of an airliner, which he  \textbf{unsuccessfully} attempted on December 22, 2001. }[...] &   Would al-Qaeda take responsibility for Richard Reid's suicide bombing attempt?      & YES & Richard Reid was a member of the Al Qaeda.   \\
 \bottomrule    
\end{tabular}
\caption{Examples of types of questions that target the implications of negated statements in \datasetname, and reasoning steps to correctly answer the questions. Negated statements are in \textcolor{blue}{blue}. Relevant categories derived from \newcite{lobue-yates-2011-types} when appropriate.}
\label{ref:qreasoning_types}
\end{table*}

\definecolor{light-gray}{gray}{0.94}
\begin{table*}[!tb]
         \centering
\footnotesize

  \begin{tabular}{p{0.1\textwidth}  p{0.8\textwidth} }
  \toprule
\textbf{Revision Strategy} & \textbf{Edited Passage} \\
  \midrule
    \multicolumn{2}{c}{\cellcolor{light-gray} \textsc{Paraphrase Edit}} \\
\multirow{4}{2cm}{\emph{Complement substitution}}  & Though Philby claimed publicly in January 1988 that he did not regret his decisions and that \sout{he missed nothing about England \textcolor{blue}{except} }\textcolor{red}{the only things he missed about England were} some friends, Colman's mustard, and Lea \& Perrins Worcestershire sauce, his wife Rufina Ivanovna Pukhova later described Philby as "disappointed in many ways" by what he found in Moscow. \\
\arrayrulecolor{black!20}\midrule
\emph{Synonym substitution}  & Local tetanus is \sout{an \textcolor{blue}{uncommon}}\textcolor{red}{a rare} form of the disease and it causes persistent contractions of muscles in the same area of the sufferer's body as where the original injury was made.\\
\arrayrulecolor{black!20}\midrule
\emph{Antonym substitution}  & The population of the Thirteen States was \sout{\textcolor{blue}{not} homogeneous}  \textcolor{red}{heterogeneous} in political views and attitudes. \\
\arrayrulecolor{black!20}\midrule
\emph{Numerical equivalence} & The period before 1920 is known as the dead-ball era, during which players would \sout{\textcolor{blue}{rarely}} hit home runs \textcolor{red}{at a low frequency}. \\
\arrayrulecolor{black!20}\midrule
\multirow{2}{*}{\emph{Ellipsis}} & \sout{Sunni scholars put trust in narrators such as Aisha, whom Shia  \textcolor{blue}{reject}}\textcolor{red}{While the Shia tend to reject} narrators such as Aisha, \textcolor{red}{Sunni scholars tend to trust them.}\\
\arrayrulecolor{black!20}\midrule
\emph{Noun-adjective conversion}  & \sout{While} Longships were used by the Norse in \sout{warfare}\textcolor{red}{a military capacity}, \sout{they were mostly used as} but mostly for \sout{troop transports}\textcolor{red}{transporting troops,} \sout{\textcolor{blue}{not}} \textcolor{red}{rather than as true} warships.\\
    \multicolumn{2}{c}{\cellcolor{light-gray} \textsc{Scope Edit}} \\
\emph{Complement inversion} &  \sout{Sunni}\textcolor{red}{Shia} scholars put trust in narrators such as Aisha, whom \sout{Shia}\textcolor{red}{Sunni} \textcolor{blue}{reject}.\\
\arrayrulecolor{black!20}\midrule
\emph{Superset-subset replacement} & During the coronavirus outbreak of 2020, alcohol sales\sout{, and even the} \textcolor{red}{were made} \textcolor{blue}{illegal}, \textcolor{red}{but the} transportation of alcohol outside of one's home\sout{, was made \textcolor{blue}{illegal}} \textcolor{red}{remained legal}.\\
\arrayrulecolor{black!20}\midrule
\emph{Attribute change}  & Moocher's look is very \sout{similar to}\textcolor{blue}{unlike} Scrooge's, except for the fact that \sout{he wears}\textcolor{red}{they both wear tattered clothes}, \sout{but \textcolor{blue}{unlike}}\textcolor{red}{and just like} his very rich cousin, Moocher is \textcolor{red}{also} a sweetheart.\\
\arrayrulecolor{black!20}\midrule
\emph{Temporal shift} & As the new Emperor \textcolor{blue}{could not} exert his constitutional powers \sout{until}\textcolor{red}{once} he came of age, a regency was set up by the National Assembly. \\
\arrayrulecolor{black!20}\midrule
\multirow{3}{*}{\emph{Veridicality}} &  \textcolor{red}{Contrary to assumptions that he was}  \textcolor{blue}{illiterate}, on arrival he was given aptitude tests which determined that \sout{he was \textcolor{blue}{illiterate}}\textcolor{red}{not only could he read the questions and respond in writing}, but \textcolor{red}{he also} had an above-average IQ of 109. \\
 
\arrayrulecolor{black}\bottomrule
  \end{tabular}

\caption{Examples of revision strategies employed by crowdworkers for paraphrase and scope edits. Categories for paraphrases are inspired by \newcite{bhagat-hovy-2013-squibs}. The negation cue is in \textcolor{blue}{blue} and newly-inserted text is in \textcolor{red}{red}. }
\label{tab:EditTypes}
\end{table*}

\begin{table*}[t]
\resizebox{\textwidth}{!}{
\begin{tabular}{p{15cm}}
\bottomrule
\textbf{Paragraph \#1:} Scorsese was initially reluctant to develop the project, though he eventually came to relate to LaMotta’s story. Schrader re-wrote Martin’s first screenplay, and Scorsese and De Niro together made uncredited contributions thereafter. \emph{Pesci was a famous actor prior to appearing in this role, but Moriarty was unknown to the producers before he suggested her for her role.} During principal photography, each of the boxing scenes was choreographed for a specific visual style and De Niro gained approximately to portray LaMotta in his later post-boxing years. Scorsese was exacting in the process of editing and mixing the film, expecting it to be his last major feature. \\
\arrayrulecolor{gray!20}\midrule
\textbf{Question:} Is it possible that the writers of this movie had specifically tailored the character to Joe Pesci’s unique on-screen charisma, with the hopes that he would accept the role? \\
\arrayrulecolor{gray!20}\midrule
\textbf{Answer:} Yes \\
\arrayrulecolor{black}\midrule
\textbf{Paragraph \#2:} Scorsese was initially reluctant to develop the project, though he eventually came to relate to LaMotta's story. Schrader re-wrote Martin's first screenplay, and Scorsese and De Niro together made uncredited contributions thereafter. \emph{Before appearing in this movie, Pesci had not achieved fame as an actor, and neither had Moriarty, who he suggested for her role.} During principal photography, each of the boxing scenes was choreographed for a specific visual style and De Niro gained approximately to portray LaMotta in his later post-boxing years. Scorsese was exacting in the process of editing and mixing the film, expecting it to be his last major feature.\\
\arrayrulecolor{gray!20}\midrule
\textbf{Question:} Is it possible that the writers of this movie had specifically tailored the character to Joe Pesci’s unique on-screen charisma, with the hopes that he would accept the role? \\
\arrayrulecolor{gray!20}\midrule
\textbf{Answer:} No \\
\arrayrulecolor{black}\bottomrule
\end{tabular}
}
\caption{Presumably, answering this question in the context of the second paragraph requires reasoning about negation, while if the question is answered in the context of the first paragraph it does not. However, if the model is only ever presented instances like the second paragraph, it is possible that there would be subtle artifacts that lead to a model's good performance without ever needing to fully process the negation.  By making minimal changes to the paragraph that intervene on the negation, we can increase our confidence that the model is able to correctly process the negation in the second paragraph.  The question-paragraph pairs must be considered jointly to accurately characterize a model's ability handle negation, hence our focus on group consistency as our preferred performance metric.}
\label{fig:appendix_interpreting_results}
\end{table*}

\section{Crowdsourcing Interface Templates}
\begin{figure*}
     \centering
     \begin{subfigure}[b]{0.8\textwidth}
         \centering
         \includegraphics[width=\textwidth]{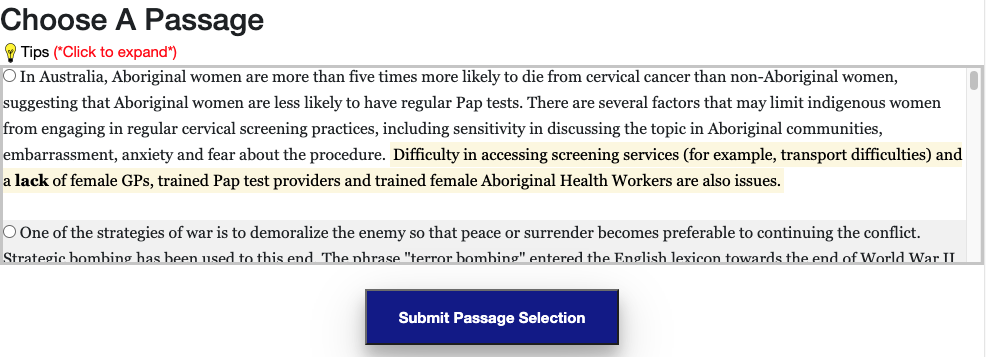}
         \caption{Crowdworkers select a passage}
     \end{subfigure}
     \hfill
     \begin{subfigure}[b]{0.8\textwidth}
         \centering
         \includegraphics[width=\textwidth]{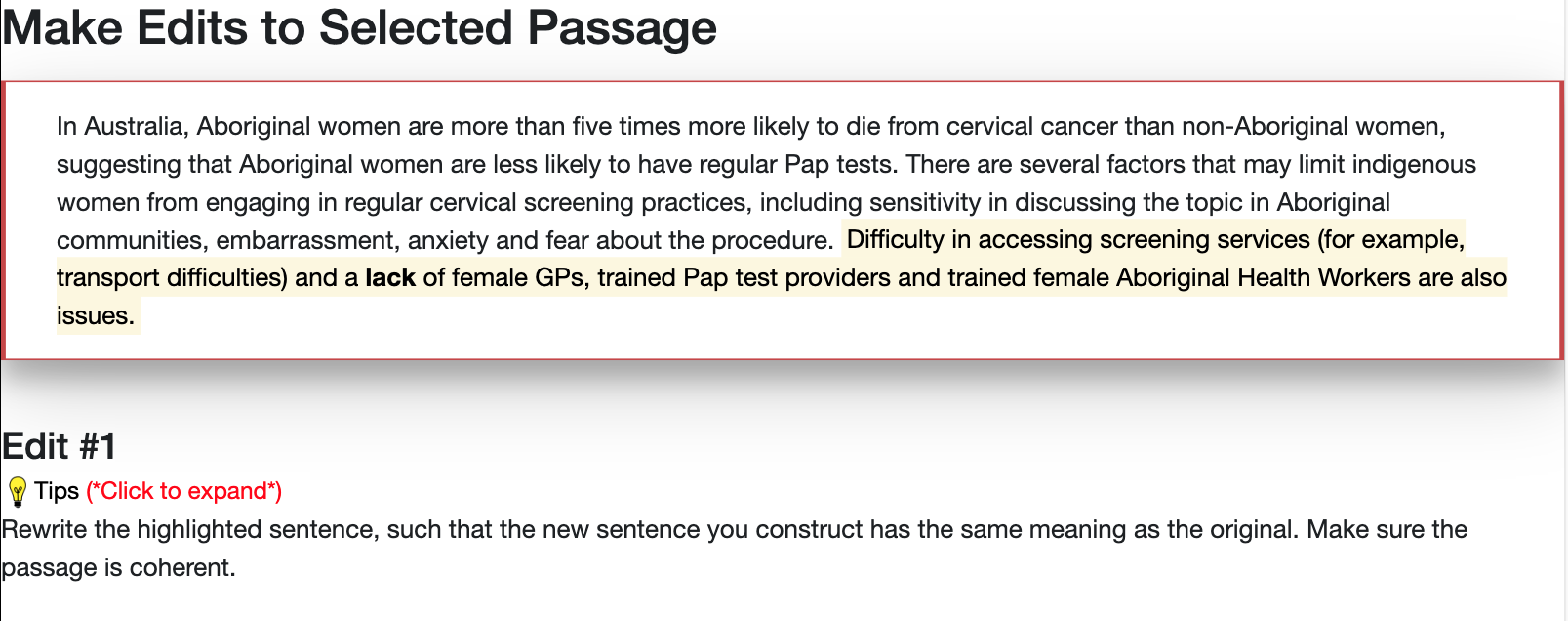}
\caption{Crowdworkers make passage edits.}
     \end{subfigure}
     \hfill
     \begin{subfigure}[b]{0.8\textwidth}
         \centering
         \includegraphics[width=\textwidth]{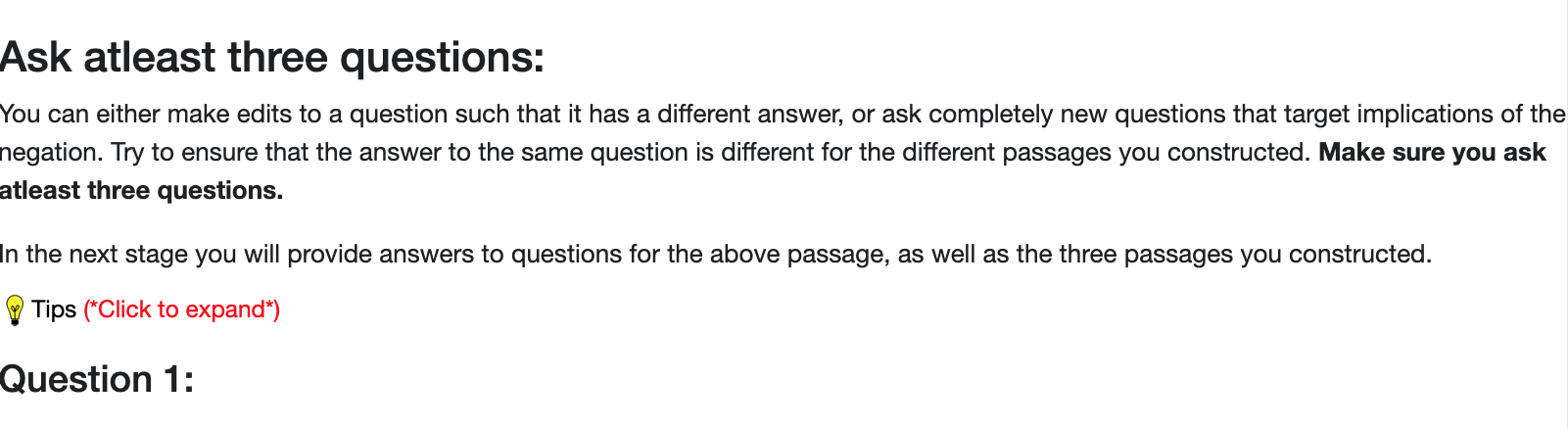}
\caption{Crowdworkers ask questions.}
     \end{subfigure}
     
          \hfill
     \begin{subfigure}[b]{0.9\textwidth}
         \centering
         \includegraphics[width=\textwidth]{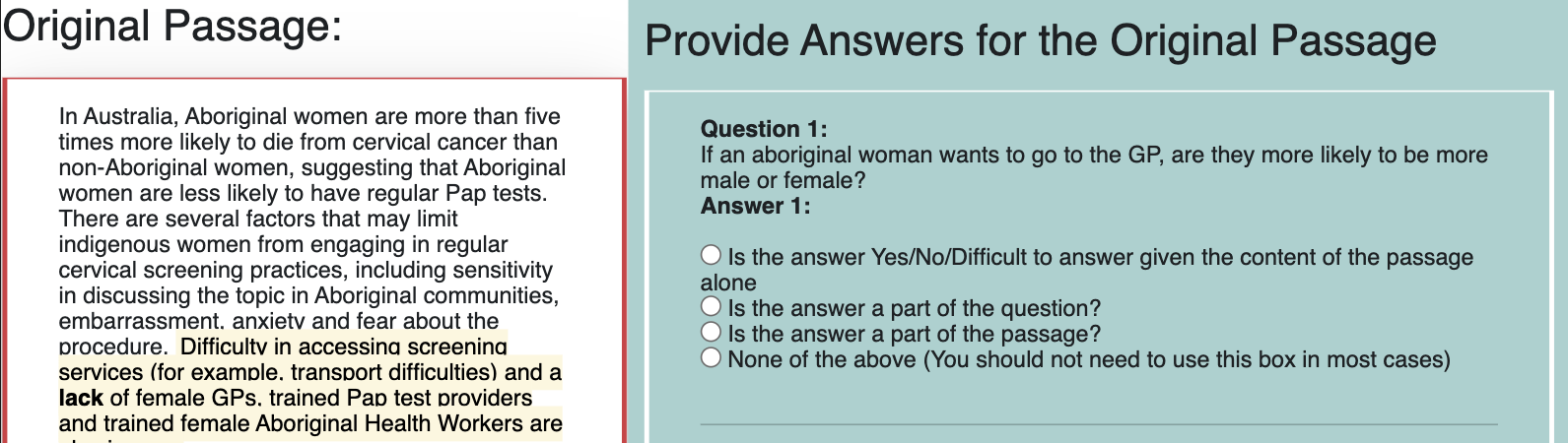}
\caption{Crowdworkers answer questions.}
     \end{subfigure}
        \caption{Sample of our Question-Answering HIT, where crowdworkers can choose a passage, make edits to that passage, ask questions about that passage and then answer those questions.}
        \label{fig:hits}
\end{figure*}

We include an example of the annotation interface we showed to crowdworkers. Figure \ref{fig:hits} shows a sample of each stage of our task.

\end{document}